\documentclass[10pt, a4paper]{article}

\usepackage{mathptmx}
\usepackage{latexsym}
\usepackage[T1]{fontenc}
\usepackage[utf8]{inputenc}
\usepackage{microtype}
\usepackage{inconsolata}
\usepackage{graphicx}
\usepackage{subcaption}
\usepackage{caption}
\usepackage{array}
\usepackage{xcolor}
\usepackage{footmisc}
\usepackage{amsmath}
\usepackage{colortbl}
\usepackage{adjustbox}
\usepackage{tabularx}
\usepackage{booktabs}
\usepackage{multirow}
\usepackage{placeins}
\usepackage[arabic,farsi,english]{babel}
\usepackage{threeparttable}
\usepackage{makecell}
\usepackage{float}
\restylefloat{table}
\restylefloat{figure}

\usepackage[final]{lrec2026} % the 'review' option anonymizes the paper following submission guideline | the 'final' option produces the camera ready version (non anonymized) | default version is 'final', so use review option for submission

\title{Can LLMs Faithfully Explain Themselves in Low-Resource Languages? A Case Study on Emotion Detection in Persian}

\name{
    Mobina Mehrazar\textsuperscript{1\ensuremath{*}},
    Mohammad Amin Yousefi\textsuperscript{1\ensuremath{*}}, \\
    \bf \large
    Parisa Abolfath Beygi\textsuperscript{2},
    Behnam Bahrak\textsuperscript{3}
}

\address{
    $^{1}$School of Electrical and Computer Engineering College of Engineering, University of Tehran, Tehran, Iran \\
    $^{2}$Computer Science Department, University of British Columbia, Vancouver, Canada \\
    $^{3}$Tehran Institute for Advanced Studies, Khatam University, Tehran, Iran \\
    m.amin.yousefi@ut.ac.ir, mobinamehrazar@ut.ac.ir, parisa.beygi@ubc.ca, b.bahrak@teias.institute
}

\abstract{
Large language models (LLMs) are increasingly used to generate self-explanations alongside their predictions, a practice that raises concerns about the faithfulness of these explanations, especially in low-resource languages. 
This study evaluates the faithfulness of LLM-generated explanations in the context of emotion classification in Persian, a low-resource language, by comparing the influential words identified by the model against those identified by human annotators. 
We assess faithfulness using confidence scores derived from token-level log-probabilities. Two prompting strategies, differing in the order of explanation and prediction (Predict-then-Explain and Explain-then-Predict), are tested for their impact on explanation faithfulness.
Our results reveal that while LLMs achieve strong classification performance, their generated explanations often diverge from faithful reasoning, showing greater agreement with each other than with human judgments. 
These results highlight the limitations of current explanation methods and metrics, emphasizing the need for more robust approaches to ensure LLM reliability in multilingual and low-resource contexts.
 \\ \newline  }

\begin{document}

\maketitleabstract

\begingroup
  \renewcommand{\thefootnote}{\ensuremath{*}} % Force the symbol to match your manual star
  \footnotetext{These authors contributed equally to this work and are listed in alphabetical order.}
\endgroup

\section{Introduction}
The widespread adoption of large language models (LLMs), both open-source and proprietary, has increased the demand for interpretability in real-world applications. A key feature of many LLMs is their ability to generate natural language explanations for predictions, known as self-explanations.

Recent studies have introduced methods for generating and evaluating self-explanations in high-resource languages like English, primarily using auto-regressive LLMs \cite{huang2023largelanguagemodelsexplain, madsen-etal-2024-self}. The quality of these explanations depends on the model's grasp of textual context, linguistic nuances, and culturally embedded meanings \cite{liu-etal-2024-multilingual}.

An open question remains: in low-resource languages, where capturing linguistic nuances is more challenging, do self-explanations accurately represent the model's decision-making? With LLMs' growing capabilities in diverse linguistic contexts, addressing this is crucial.

To explore this, we evaluate the faithfulness of self-explanations in Persian, focusing on emotion detection. Using a Persian emotion-labeled corpus with formal and informal samples, we assess challenges in both language understanding and explanation generation.

Our analysis uses auto-regressive LLMs, which identify keywords from the input as explanations for their predictions. We adopt two paradigms: Predict-then-Explain (P-E) and Explain-then-Predict (E-P) \cite{NIPS2018_8163}, to explore the relationship between explanations and predictions. By isolating influential words, we assess the model's reasoning by measuring changes in predictions and confidence when these words are added or removed.

Previous work has often relied on self-reported confidence, where models express certainty through natural language explanations or specific words \cite{huang2023largelanguagemodelsexplain}. 
However, empirical evidence shows that self-reported confidence estimates often diverge from calibrated confidence estimates, limiting their reliability for faithfulness evaluation \cite{zhou-etal-2024-relying}.

To overcome this limitation, we utilize token-level log-probabilities obtained through API access to compute more grounded confidence estimates \cite{xue2024comprehensivestudymultilingualconfidence}. 
Throughout our analysis, we observe a consistent tendency toward overconfident predictions, which can distort faithfulness assessments. Accordingly, we apply temperature scaling \citep{10.5555/3305381.3305518}, a post hoc calibration technique to improve the reliability of confidence estimates derived from token-level log-probabilities. These calibrated scores are then used to evaluate the faithfulness of self-explanations through a set of quantitative metrics.

While recent studies have explored the alignment between human reasoning and model-generated explanations \citep{venkatesh2024comparing}, research is limited for low-resource languages. To address this, we investigate whether self-explanations in Persian align with human reasoning, a key step in evaluating LLM interpretability and reliability across diverse linguistic and cultural contexts. We designed a two-stage human annotation process that mirrors the output format of the models, enabling direct and fair comparisons with their explanations.

\section{Related Work}
\subsection{Feature Attribution Methods}
Recent work in explainable AI has focused on enhancing the interpretability of LLMs through feature attribution, which identifies input elements that most influence model outputs. Two main approaches have emerged. One involves analyzing the model's sensitivity to small perturbations in input features, including gradient-based methods \citep{kindermans2016investigatinginfluencenoisedistractors, 10.5555/3305890.3306006}, which trace relevance through internal model mechanisms. These methods are considered model-specific or analytical explanations. However, they typically require access to model weights, attention maps, or input gradients, assumptions that do not hold for closed-source LLMs, making them inapplicable in our setting.

Another approach to defining feature importance in LLMs is perturbation-based analysis, where input tokens, identified by model explanations, are altered to assess their impact on model predictions. Methods like LIME \citep{10.1145/2939672.2939778} and SHAP \citep{SHAP_NIPS2017} capture non-linear interactions by perturbing multiple tokens and assigning importance through linear regression (LIME) or Shapley values (SHAP). Recent adaptations for LLMs include TokenSHAP for Shapley-based token attribution \citep{horovicz-goldshmidt-2024-tokenshap} and studies on LIME's explainability across model sizes \citep{heyen2024the}.

We adopt a model-compatible approach to feature attribution by prompting the LLM to generate self-explanations, identifying the most influential tokens for its prediction. Unlike gradient- or attention-based methods, this strategy directly elicits the model's reasoning in natural language or extractive form. We evaluated the quality of these self-explanations by masking the identified tokens and measuring their impact on predictions. This work extends recent studies, as discussed in the following.

\subsection{Self-Explanations via Input Token Attribution in LLMs}

\citet{10.1007/978-3-031-78977-9_3} examines whether lightweight LLMs (up to 8B parameters, like Gemma and Llama3-8B) can reliably generate self-explanations as influential words or phrases. The study, focusing on (P-E) paradigm, compares these explanations primarily with gradient-based methods. Although tasks like food hazard classification and sentiment analysis are explored, the lack of benchmarking on more widely-used models like GPT limits the generalizability to high-performing, real-world LLMs.

\citet{huang2023largelanguagemodelsexplain} prompt ChatGPT to identify the top-\textit{k} most important tokens for binary sentiment classification and evaluate the faithfulness of these self-explanations using metrics like \emph{comprehensiveness}, \emph{sufficiency}, and \emph{decision flip rate} (DFMIT). They compare self-explanations with traditional methods like LIME \cite{10.1145/3375627.3375830} and occlusion \cite{li2017understandingneuralnetworksrepresentation}, exploring both E-P and P-E paradigms. Notably, their evaluation relies on confidence scores generated by the model itself, a potentially unreliable method not based on internal probabilities.

\citet{venkatesh2024comparing} offers a complementary perspective by using eye-tracking to compare human, ML, and LLM-derived explanations in text classification. They extract top words from each source, LLM attributions, attention-based models, and human gaze saliency, to measure alignment. Unlike our method, they rely on implicit gaze behavior rather than explicitly asking humans to identify influential tokens, marking a key methodological distinction.

\citet{madsen-etal-2024-self} uses a redaction-based consistency framework to assess explanation reliability, focusing on prediction flips. While it explores counterfactual explanations, it lacks models like GPT and human annotations, distinguishing it from our study.

\subsection{Research Gap}
To the best of our knowledge, there is no existing work that performs confidence estimation using the model's log probabilities for evaluating self-explanations and subsequently calibrating these estimates to improve explanation quality. Furthermore, research on emotion detection in Persian, a resource-limited language, is lacking, which this study aims to address.

\section{Methodology}
This section outlines the experimental design used to evaluate the faithfulness of self-explanations generated by LLMs for emotion classification in Persian. We detail our interaction protocol with LLMs, the explanation extraction strategies employed, the approach to estimating model confidence and calibrating confidence, the structure of the human annotation process, and the evaluation metrics used to assess explanation faithfulness and human agreement.

\subsection{Interaction Protocol for Auto-regressive LLMs}
Auto-regressive LLMs generate text sequentially, predicting one token at a time based on prior context. This left-to-right process enhances coherence and context awareness, making them well-suited for natural language generation. Built on transformer architectures, these models use self-attention to capture long-range dependencies.

In our experiments, we accessed each LLM through its API using a standardized prompting protocol with three distinct roles:

\begin{itemize}
\item \texttt{System:} specifies high-level instructions that govern the behavior of the model across all tasks.
\item \texttt{User:} provides task-specific queries, prompts, or instructions.
\item \texttt{Assistant:} generates responses, including classifications and explanations, based on the given instructions and queries.
\end{itemize}

This structured protocol ensures consistency in model behavior and output formatting throughout all stages of evaluation.

\subsection{Explanation Extraction Approaches for the Classification Task}
We use LLMs for emotion classification on a Persian-labeled corpus and for generating self-explanations that highlight influential words behind each prediction. To examine how the order of explanation and prediction impacts model behavior and explanation quality, we adopt two prompting paradigms from \citet{NIPS2018_8163}:

\begin{itemize}
\item \textbf{Explain-then-Predict} (E-P): The model first generates an explanation from the input text, then predicts the emotion label using that explanation as context.
\item \textbf{Predict-then-Explain} (P-E): The model first predicts the emotion label from the input, then explains its decision based on the classification and prior dialogue.
\end{itemize}
Under both paradigms, the model classifies the text into one of six emotion categories: Sadness (0), Happiness (1), Anger (2), Surprise (3), Hatred (4), and Fear (5). The model outputs only the corresponding numeric label. Furthermore, the explanations are constrained to a set of the top-\textit{k} most influential words, all of which must be present in the original Persian input to ensure textual alignment.

In the E-P setting, influential words are extracted first and used to condition the classification. In the P-E setting, the model first classifies the full text, then generates an explanation highlighting the top influential words.

To evaluate the faithfulness of the generated explanations, we prompt the model with only the extracted influential words for emotion classification. If the model predicts the same label with high confidence, it suggests the explanation accurately reflects the decision rationale.

We also replace the top influential words with a placeholder token and reclassify the modified text. A drop in classification confidence or a change in prediction indicates the importance of the removed words.

These changes in the class and confidence of predictions will be evaluated through the \emph{faithfulness} evaluation metrics, further discussed in \ref{subsec:evaluationMetrics}.

\subsection{Confidence Estimation and Calibration}
Evaluating the faithfulness of self-generated explanations requires reliable confidence estimates in the model's predictions. We use token-level log-probabilities of predicted labels, transforming them into normalized confidence scores (0 to 1).

To address the overconfidence of LLMs' probability estimates, we apply temperature scaling \citep{10.5555/3305381.3305518}, a post-hoc calibration method. This introduces a scalar parameter, \emph{Temperature}, optimized on a held-out calibration set by minimizing the \emph{Expected Calibration Error} (ECE). The optimal \emph{temperature} is then applied to the model's output probabilities during evaluation, improving the reliability of confidence estimates for faithfulness assessment.

The ECE is computed as:
\begin{equation} \text{ECE} = \sum_{m=1}^M \frac{|B_m|}{n} \left| \text{acc}(B_m) - \text{conf}(B_m) \right| \end{equation}

where $B_m$ denotes the set of predictions whose confidence scores fall into the $m$-\textit{th} bin. To construct the bins, the confidence interval $[0,1]$ is partitioned into $M$ equal-width sub-intervals 
\[
I_m = \Big[ \tfrac{m-1}{M}, \tfrac{m}{M} \Big), \quad m = 1, \ldots, M,
\]
and each prediction with confidence $\hat{p}_i$ is assigned to bin $B_m$ if $\hat{p}_i \in I_m$. Here, $n$ is the total number of samples, $\text{acc}(B_m)$ is the average accuracy of bin $B_m$, and $\text{conf}(B_m)$ is the average confidence in that bin.

\begin{table*}[h]
  \centering
  \small
  \renewcommand{\arraystretch}{1.2}
  \setlength{\tabcolsep}{4pt}
  \begin{tabular}{c|cccc|cccc}
    \hline
    \textbf{Model} & \multicolumn{4}{c|}{\textbf{Predict-then-Explain (P-E)}} & \multicolumn{4}{c}{\textbf{Explain-then-Predict (E-P)}} \\
    & \textbf{Precision} & \textbf{Recall} & \textbf{F1-Score} & \textbf{Accuracy} & \textbf{Precision} & \textbf{Recall} & \textbf{F1-Score} & \textbf{Accuracy} \\
    \hline
    \textbf{GPT-4o}      & \textbf{81} & \textbf{78} & \textbf{77} & \textbf{77} & \textbf{78} & \textbf{75} & \textbf{74} & \textbf{75} \\
    \textbf{GPT-4-turbo} & 75 & 72 & 70 & 72 & 74 & 68 & 67 & 68 \\
    \textbf{Llama3.3-70B-instruct} & 74 & 65 & 61 & 65 & 79 & 62 & 59 & 62 \\
    \textbf{DeepSeek-V3} & 58 & 65 & 60 & 65 & 71 & 62 & 59 & 62 \\
    \hline
    \textbf{Human Annotators} & 69 & 66 & 67 & 66 & 69 & 66 & 67 & 66 \\
    \hline
  \end{tabular}
  \captionsetup{width=\linewidth}
  \caption{Comparison of classification performance in P-E and E-P settings. All metrics are reported as percentages.}
  \label{tab:performance_metrics}
\end{table*}

\subsection{Human Annotation Process}
To allow for a fair comparison with the model, we implemented a two-stage human annotation process that closely mirrors the structure of the LLM-generated outputs.

Twenty-five native Persian speakers participated as annotators, providing emotion classifications and influential words in a format consistent with the model's output. Unlike the model, annotators selected the top-\textit{k} influential words and classified texts without a specific order, reflecting human decision-making. We did not require adherence to the E-P or P-E paradigms or collect explicit confidence scores, as these are not naturally expressed in human annotation.

In the first stage, each annotator received a distinct subset of full-text samples, classifying each into one of six emotion categories and identifying the top-\textit{k} influential words guiding their decision.

In the second stage, annotators classified two modified versions of previously unseen samples: (1) one with only the top-\textit{k} influential words, and (2) one with the top-\textit{k} words replaced by a placeholder token. Each version was classified into one of the six emotion categories.

To prevent familiarity bias and ensure decisions were based solely on the presented text, each annotator was assigned a unique set of samples in both stages. Additionally, all annotations underwent secondary verification, where they were cross-checked and validated by other annotators for consistency and correctness.

\subsection{Evaluation Metrics}
\label{subsec:evaluationMetrics}
To evaluate the faithfulness of explanations generated by the underlying LLMs and human annotators, we employ a set of established evaluation metrics. In addition, we measure the pairwise agreement between model- and human-generated explanations. The metrics used in this study are detailed below.

\subsubsection{Faithfulness Metrics}
\textbf{Comprehensiveness} measures the impact of the most influential words on a model's prediction by removing them and observing the change in confidence \citep{deyoung-etal-2020-eraser}. The score is the difference in confidence before and after removal, with a higher score indicating a more faithful explanation.

\textbf{Sufficiency} measures whether the most influential words alone can drive the model's prediction with the same confidence \citep{deyoung-etal-2020-eraser}. It is calculated as the difference in confidence between the full-text input and the input containing only the influential words. A lower score indicates higher \emph{sufficiency}, meaning the influential words alone are nearly sufficient for the prediction.

\textbf{Decision Flip Rate (Top-\textit{k} Only)} measures the model's reliance on broader context by quantifying how often its prediction changes when only the top-\textit{k} influential words are retained. Noted as DF$_\text{TopKOnly}$, it averages the prediction flips across all samples. Higher values indicate that the selected words alone are insufficient to reproduce the original decision, suggesting a greater need for contextual information.

\textbf{Decision Flip Rate (Top-\textit{k} Removed)} quantifies how often a model's prediction changes when its most influential words are replaced with placeholders \cite{chrysostomou-aletras-2021-improving}. Noted as DF$_\text{TopKRemoved}$, it calculates the average rate of prediction flips across the dataset, reflecting the model's sensitivity to key input words. Higher values indicate greater reliance on these words, highlighting their critical role in decision-making.

\subsubsection{Agreement Metrics}

\textbf{Feature Agreement} measures the overlap between the top-\textit{k} most influential features identified by two different explanation methods. It is calculated as the fraction of common features between the two sets of top-\textit{k} features. Higher values indicate greater alignment in what both sources consider important, suggesting stronger agreement in explanatory focus \citep{krishna2024the}.

\textbf{Intersection-over-Union (IoU)} measures the similarity between two sets of top-\textit{k} influential features by comparing their intersection relative to their union. It is defined as 
\[
\text{IoU}(A,B) = \frac{|A \cap B|}{|A \cup B|},
\]
where $A$ and $B$ are the feature sets under comparison. Values range from 0 (no overlap) to 1 (perfect overlap). Higher values indicate stronger agreement between the two explanation methods in terms of the features they identify as important.

\section{Experiments and Results}
\subsection{Experimental Setup}

Following our methodology, we selected a set of LLMs that offer access to token-level log-probabilities via their official APIs. These models vary in architecture and scale, with parameter counts ranging from approximately 70 billion to over one trillion. Specifically, our experiments used Llama3.3-70B-instruct \cite{grattafiori2024llama3herdmodels} (70B parameters), DeepSeek-V3 \cite{deepseekai2025deepseekv3technicalreport} (671B parameters), GPT-4-turbo \cite{openai2024gpt4technicalreport}, and GPT-4o \cite{openai2024gpt4ocard}, both estimated to have over one trillion parameters.

We conducted our experiments on a subset of the ARMANEMO dataset \citep{mirzaee2025armanemo}, a manually annotated Persian corpus for emotion detection in both formal and colloquial text. ARMANEMO comprises over 7,000 sentences labeled with six emotion categories and an additional OTHER class for out-of-scope emotions. To ensure focused evaluation, we excluded Other-labeled samples and curated a balanced evaluation set of 300 instances (about 50 per emotion). An additional disjoint set of 210 samples (35 per emotion) was used for temperature scaling to calibrate model confidence.

To select an appropriate value for $k$, we conducted a small-scale empirical analysis on a subset of the data. While dynamic values based on text length were initially considered, they often underperformed: many short texts contained multiple influential words, and low $k$ values failed to capture them all. To address this, we chose a fixed $k=5$ for all samples. Although this may introduce less informative words in shorter texts, it ensures emotionally salient words are not missed, which is crucial for generating faithful and comprehensive explanations.

\subsection{Emotion Classification Performance}{\label{sec:EmotionClassificationPerformance}}

\begin{table*}[h]
\centering
\small
\renewcommand{\arraystretch}{1.2}
\setlength{\tabcolsep}{3.5pt}
\begin{tabular}{llcccccccccccc}
\toprule
\textbf{Model} & \textbf{Metric} & \multicolumn{2}{c}{\textbf{Sadness}} & \multicolumn{2}{c}{\textbf{Happiness}} & \multicolumn{2}{c}{\textbf{Anger}} & \multicolumn{2}{c}{\textbf{Surprise}} & \multicolumn{2}{c}{\textbf{Hatred}} & \multicolumn{2}{c}{\textbf{Fear}} \\
\cmidrule(lr){3-4} \cmidrule(lr){5-6} \cmidrule(lr){7-8} \cmidrule(lr){9-10} \cmidrule(lr){11-12} \cmidrule(lr){13-14}
& & P-E & E-P & P-E & E-P & P-E & E-P & P-E & E-P & P-E & E-P & P-E & E-P \\
\midrule
\multirow{3}{*}{\textbf{GPT-4o}} 
& Precision & 79 & 70 & 94 & 96 & 58 & 55 & 92 & 95 & 90 & 70 & 74 & 81 \\
& Recall    & 63 & 71 & 90 & 96 & 92 & 76 & 46 & 40 & 75 & 78 & 100 & 88 \\
& F1-Score  & 70 & 70 & 92 & 96 & 71 & 64 & 61 & 56 & 82 & 74 & 77 & 84 \\
\midrule
\multirow{3}{*}{\textbf{GPT-4-turbo}} 
& Precision & 81 & 74 & 90 & 89 & 51 & 42 & 81 & 94 & 79 & 55 & 69 & 87 \\
& Recall    & 25 & 54 & 86 & 96 & 82 & 49 & 60 & 34 & 86 & 92 & 94 & 85 \\
& F1-Score  & 38 & 62 & 88 & 92 & 63 & 45 & 69 & 50 & 82 & 69 & 80 & 86 \\
\midrule
\multirow{3}{*}{\textbf{Llama3.3-70B-instruct}} 
& Precision & 68 & 79 & 45 & 100 & 65 & 89 & 69 & 77 & 37 & 100 & 100 & 94 \\
& Recall    & 69 & 96 & 90 & 04 & 63 & 71 & 67 & 94 & 100 & 10 & 35 & 65 \\
& F1-Score  & 69 & 86 & 60 & 08 & 64 & 79 & 68 & 85 & 54 & 18 & 52 & 77 \\
\midrule
\multirow{3}{*}{\textbf{DeepSeek-V3}} 
& Precision & 66 & 75 & 79 & 83 & 43 & 35 & 00 & 100 & 71 & 48 & 91 & 87 \\
& Recall    & 73 & 58 & 96 & 96 & 78 & 43 & 00 & 08 & 82 & 88 & 62 & 81 \\
& F1-Score  & 69 & 65 & 86 & 89 & 55 & 39 & 00 & 15 & 76 & 62 & 74 & 84 \\
\midrule
\multirow{3}{*}{\textbf{Human Annotators}} 
& Precision & \multicolumn{2}{c}{62} & \multicolumn{2}{c}{95} & \multicolumn{2}{c}{50} & \multicolumn{2}{c}{71} & \multicolumn{2}{c}{55} & \multicolumn{2}{c}{82} \\
& Recall    & \multicolumn{2}{c}{63} & \multicolumn{2}{c}{82} & \multicolumn{2}{c}{67} & \multicolumn{2}{c}{54} & \multicolumn{2}{c}{67} & \multicolumn{2}{c}{65} \\
& F1-Score  & \multicolumn{2}{c}{63} & \multicolumn{2}{c}{88} & \multicolumn{2}{c}{57} & \multicolumn{2}{c}{61} & \multicolumn{2}{c}{60} & \multicolumn{2}{c}{72} \\
\bottomrule
\end{tabular}
\captionsetup{width=\linewidth}
\caption{Comparison of classification Performance across P-E and E-P paradigms for each emotion separately on full-text samples. All metrics are reported as percentages.}
\label{tab:performance_metrics_per_class}
\end{table*}

Table~\ref{tab:performance_metrics} reports macro-level classification performance across the E-P and P-E prompting paradigms. Consistently, P-E outperforms E-P across all models, suggesting that predicting before explaining leads to better results, echoing findings by \citeauthor{huang2023largelanguagemodelsexplain} and \citeauthor{NIPS2018_8163}.

As shown, ChatGPT-family models outperform even human annotators, particularly in terms of accuracy and F1-score. In contrast, Llama3.3-70B-instruct and DeepSeek-V3 underperform relative to humans with a slight difference, highlighting current limitations in their ability to interpret Persian emotional content comparing to ChatGPT-family models. The comparatively lower performance of human annotators does not necessarily indicate inferior emotional interpretation but rather highlights a fundamental characteristic of the evaluation benchmark. Many texts in the dataset express a mix of emotions, where the 'gold' label represents a majority vote. An LLM, as a probabilistic model, is optimized to predict the most statistically likely outcome, i.e. the majority label. In contrast, a human annotator might legitimately identify a valid but secondary emotion and be marked as 'incorrect' by a benchmark that only accepts a single answer. Thus, the models' superior scores may reflect their alignment with the benchmark's majoritarian nature more than a 'truer' understanding of the emotional content.

% The comparatively lower performance of human annotators is understandable, as many texts in the dataset express multiple emotions, but only the label chosen by the majority is retained, leading to potential disagreement with our annotators' choices.

Table~\ref{tab:performance_metrics_per_class} provides a detailed comparison of model performance by emotion category under the E-P and P-E settings. GPT-family models show relatively stable results across both paradigms, indicating robustness in handling subtle or context-dependent emotions. Generally, E-P yields higher F1-scores for Happiness, Fear, and Sadness, while P-E performs better for Anger and Hatred. The Surprise category poses consistent challenges across models, likely due to its nuanced and context-sensitive expression in Persian, often involving tone, sarcasm, or ambiguity. Misclassifications of Surprise frequently map to Anger or Happiness. Notably, DeepSeek-V3 fails to correctly classify any samples from the Surprise category in the E-P setup, highlighting the difficulty of detecting this emotion class. Additionally, a high degree of confusion between Anger and Hatred suggests substantial semantic and affective overlap, complicating their disambiguation.

\subsection{Process for Collecting Explanations} 
To collect self-explanations, both language models and human annotators were asked to identify the top 5 most influential words contributing to their emotion classification. All selected words had to exactly match those in the original input.

\subsection{Improving Confidence Estimation}\label{subsec:ImprovingConfidenceEstimation}
To improve the reliability of confidence estimates for explanation faithfulness, we applied temperature scaling independently for each model and prompting paradigm. A grid search over the range [0.1, 21.0] with a 0.1 step size identified the optimal \emph{temperature} on a calibration set, which was then applied to the evaluation set.

\begin{table}[h!]
    \centering
    \small
    \renewcommand{\arraystretch}{1.4}
    \setlength{\tabcolsep}{3.5pt}
    \begin{tabular}{llcc}
        \toprule
        \textbf{Model} & \textbf{Dataset \& Scaling} & \multicolumn{2}{c}{\textbf{ECE}} \\
        \cmidrule(lr){3-4}
        & & \textbf{P-E} & \textbf{E-P} \\
        \midrule
                        & Pre-Scale Calib. Set & 23.26 & 25.04 \\
        \textbf{GPT-4o} & Post-Scale Calib. Set & 6.20 & 6.68 \\
                        & Post-Scale Eval. Set & 4.83 & 6.64 \\
        \midrule
                            & Pre-Scale Calib. Set & 36.96 & 37.22 \\
       \textbf{GPT-4-turbo} & Post-Scale Calib. Set & 6.90 & 3.57 \\
                            & Post-Scale Eval. Set & 4.99 & 11.52 \\
        \midrule
        \multirow{3}{*}{\parbox{2cm}{\textbf{Llama3.3-\\70B-instruct}}}
                            & Pre-Scale Calib. Set & 43.19 & 47.31 \\
                            & Post-Scale Calib. Set & 2.36 & 3.57 \\
                            & Post-Scale Eval. Set & 1.40 & 4.15 \\
        \midrule
                             & Pre-Scale Calib. Set & 39.26 & 37.40 \\
        \textbf{DeepSeek-V3} & Post-Scale Calib. Set & 7.73 & 5.78 \\
                             & Post-Scale Eval. Set & 5.96 & 5.57 \\
        \bottomrule
    \end{tabular}
    \captionsetup{width=\linewidth}
    \caption{Temperature optimization and calibration results across models, reported as percentages.}
    \label{tab:calibrationResults}
\end{table}

Table~\ref{tab:calibrationResults} shows that temperature scaling significantly reduces ECE across all models and prompting paradigms. These improvements generalize well to the evaluation set, indicating that the calibrated temperatures effectively correct the models' overconfident predictions.

\subsection{Faithfulness Evaluation}

\begin{table*}[t!]
    \centering
    \small
    \renewcommand{\arraystretch}{1.2}
    \setlength{\tabcolsep}{4.7pt}
    \begin{tabular}{lcccccc}
        \toprule
        \textbf{Model} & \textbf{Comp~(↑)} & \textbf{Suff~(↓)} & \textbf{DF\textsubscript{TopKRemoved}~(↑)} & \textbf{DF\textsubscript{TopKOnly}~(↓)} \\
        \midrule
        \multicolumn{5}{c}{\textbf{Predict-then-Explain (P-E)}} \\
        \midrule
        \textbf{GPT-4o} & \textbf{27.42} & 8.08 & 38.00 & \textbf{22.00} \\
        \textbf{GPT-4-turbo} & 17.76 & \textbf{-1.57} & \textbf{53.33} & 24.33 \\
        \textbf{Llama3.3-70B-instruct} & 7.96 & 7.84 & 30.33 & 29.33 \\
        \textbf{DeepSeek-V3} & 18.80 & 12.74 & 39.67 & 29.00 \\
        \midrule
        \multicolumn{5}{c}{\textbf{Explain-then-Predict (E-P)}} \\
        \midrule
        \textbf{GPT-4o} & 22.82 & 24.76 & 31.67 & 36.00 \\
        \textbf{GPT-4-turbo} & 24.36 & 24.56 & 43.00 & 37.67 \\
        \textbf{Llama3.3-70B-instruct} & 12.22 & \textbf{18.59} & 27.00 & 42.00 \\
        \textbf{DeepSeek-V3} & \textbf{28.77} & 22.08 & \textbf{48.33} & \textbf{30.00} \\
        \midrule
        \textbf{Human Annotators} & - & - & 53.33 & 39.00 \\
        \bottomrule
    \end{tabular}
    \captionsetup{width=\linewidth}
    \caption{Faithfulness comparison of underlying LLMs in P-E and E-P settings, after confidence calibration. All metrics are reported as percentages.}
    \label{tab:faithfulness_comparison_after_calibration}
\end{table*}

We assess how explanation sequencing affects the faithfulness of self-generated explanations by comparing the E-P and P-E prompting paradigms across different LLMs.

Faithfulness is evaluated under both pre- and post-calibration settings. As discussed in Section~\ref{subsec:ImprovingConfidenceEstimation}, calibration improves the reliability of confidence scores, notably enhancing \emph{Sufficiency} in the P-E setting. This effect is primarily due to temperature scaling reducing overall confidence values, with higher-confidence predictions being scaled down more aggressively.

Since predictions based on the full-text input typically have higher confidence than those based on either the top-\textit{k} words only or the top-\textit{k} words removed, calibration narrows the confidence gap between these modes. 
As a result, \emph{Sufficiency}, which measures how well the top-\textit{k} words alone preserve the model's confidence, tends to increase. However, the same compression in confidence scores reduces the gap between the full-text input and the top-\textit{k} words removed input, leading to lower \emph{Comprehensiveness}.

As shown in Table~\ref{tab:faithfulness_comparison_after_calibration}, most models exhibit greater explanation faithfulness under the P-E setup. This aligns with prior findings on English-language datasets \citep{huang2023largelanguagemodelsexplain}, suggesting that generating explanations after predictions yields more faithful results, even for Persian emotion classification. DeepSeek-V3 is a notable exception, performing better when influential words are identified before prediction. However, Llama3.3-70B-instruct consistently shows weaker faithfulness across both paradigms, with a notable decline in \emph{comprehensiveness} after calibration, possibly due to its significantly smaller size compared to the other models.

In the P-E setup, While no single model outperforms across all metrics, clear patterns emerge: GPT-4o leads in \emph{comprehensiveness} and DF\textsubscript{TopKOnly}, whereas GPT-4-turbo excels in \emph{sufficiency} and DF\textsubscript{TopKRemoved}. These results suggest that both models are more effective in identifying the most influential words in their predictions. But Unexpectedly, in the E-P setting, DeepSeek-V3 outperformes other models in most metrics. 

Interestingly, and contrary to expectations, GPT-4-turbo produces more confident predictions when given only the identified influential words, without additional context.

Compared to the models, human annotators perform well on DF\textsubscript{TopKRemoved}, indicating that their emotion predictions rely heavily on the identified influential words, and removing these often changes their decisions.

\subsection{Agreement Evaluation between LLM and Human Reasoning}

\begin{figure*}[t]
\centering
\begin{subfigure}[t]{0.8\textwidth}
    \centering
    \begin{subfigure}[t]{0.48\textwidth}
        \includegraphics[width=\textwidth, height=\textwidth]{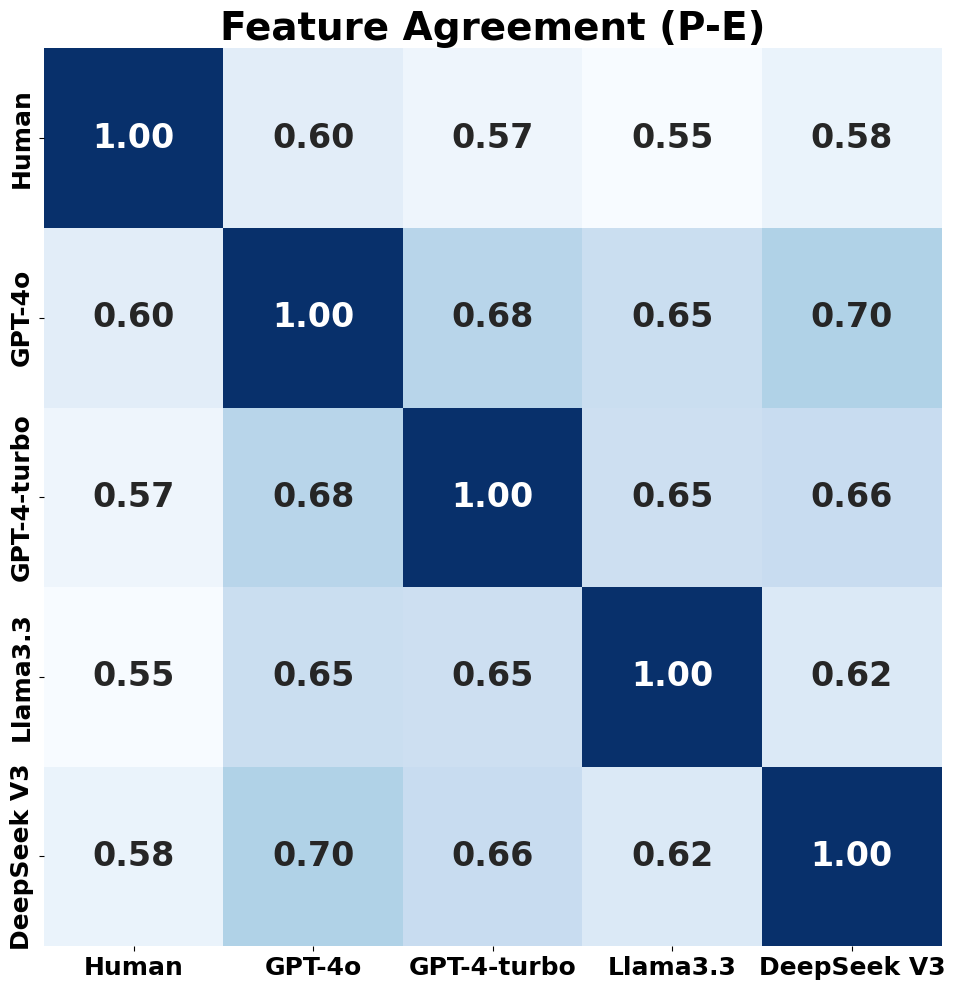}
    \end{subfigure}
    \hfill
    \begin{subfigure}[t]{0.48\textwidth}
        \includegraphics[width=\textwidth, height=\textwidth]{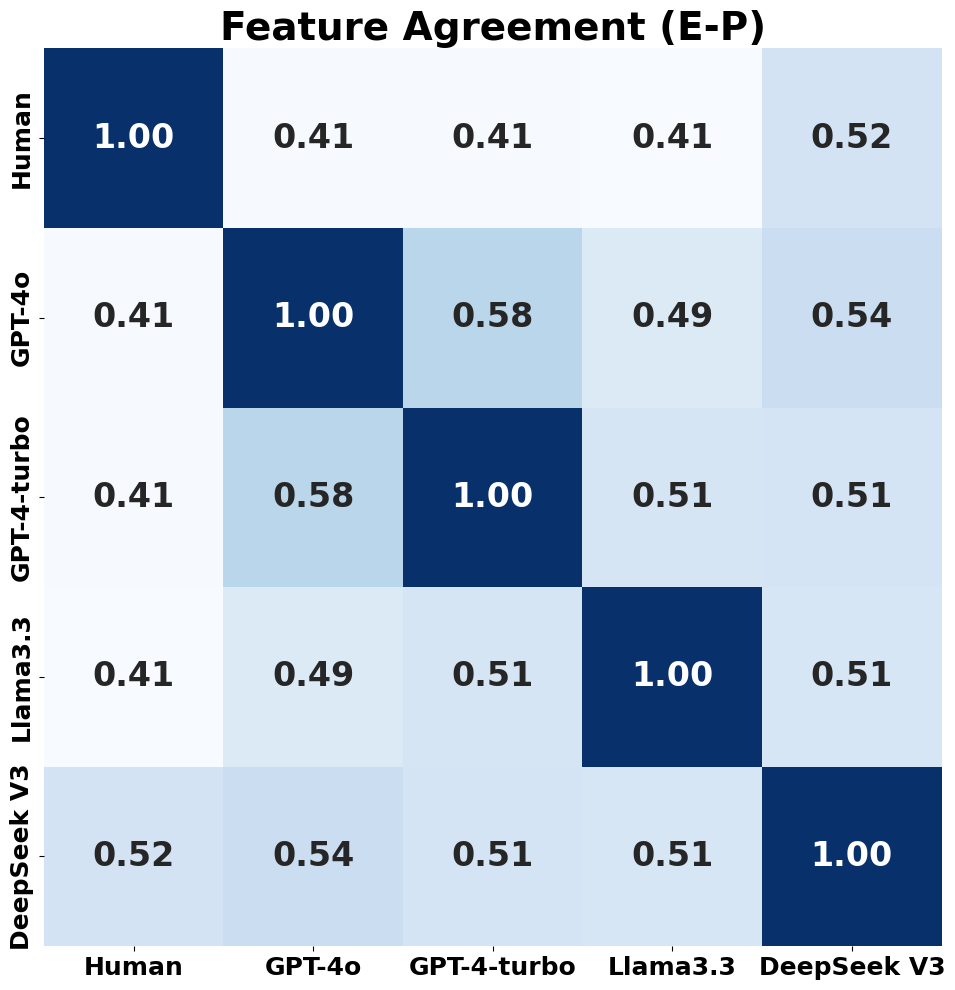}
    \end{subfigure}
\end{subfigure}
\hfill
\begin{subfigure}[t]{0.8\textwidth}
    \centering
    \begin{subfigure}[t]{0.48\textwidth}
        \includegraphics[width=\textwidth, height=\textwidth]{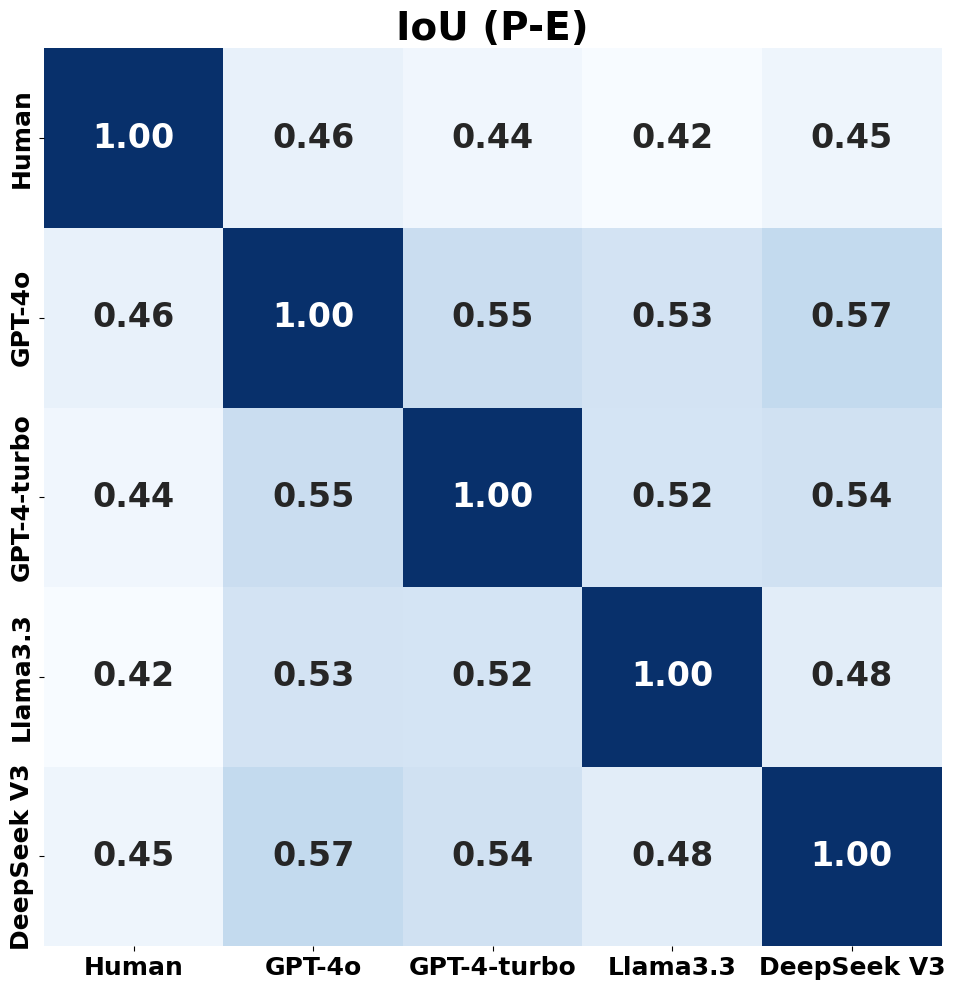}
    \end{subfigure}
    \hfill
    \begin{subfigure}[t]{0.48\textwidth}
        \includegraphics[width=\textwidth, height=\textwidth]{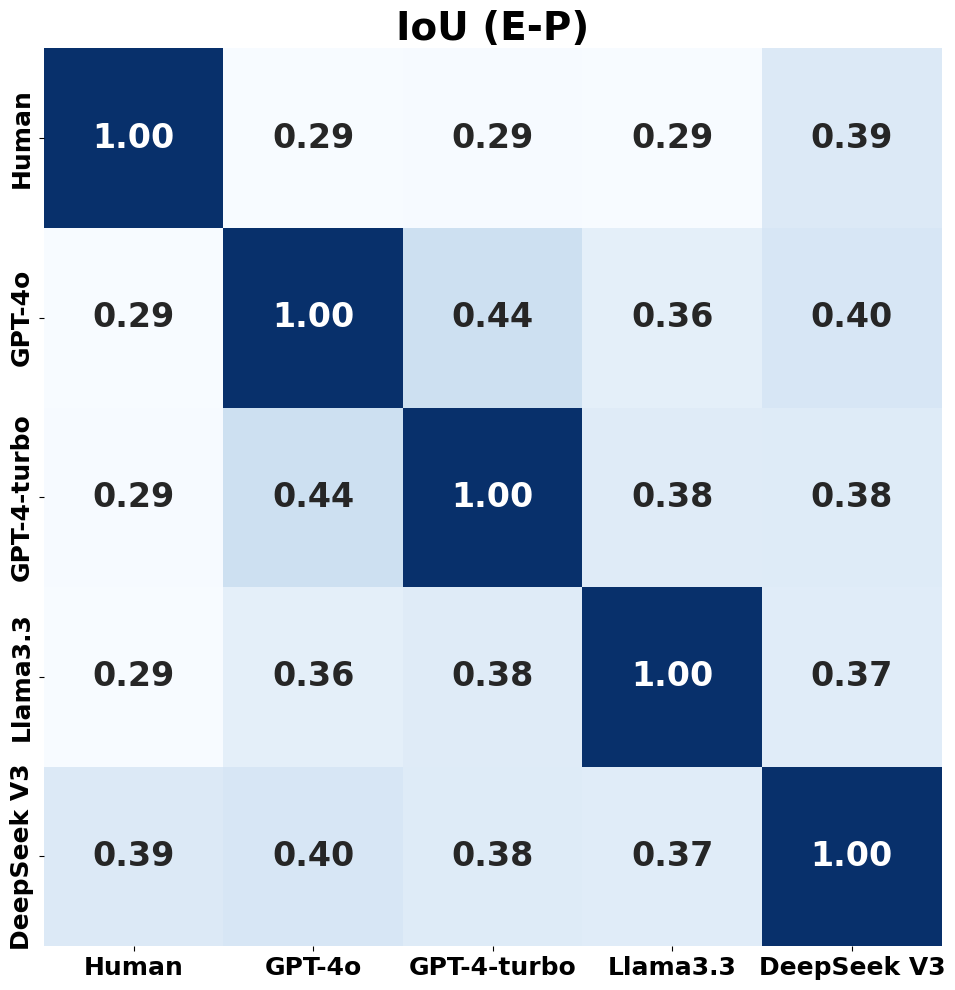}
    \end{subfigure}
\end{subfigure}
\captionsetup{width=\linewidth}
\caption{Comparison of explanation agreement metrics between human annotators and model explanations under both P-E and E-P prompting paradigms.}
\label{fig:agreement}
\end{figure*}

Figure~\ref{fig:agreement} shows the agreement between model-generated and human-annotated top-\textit{k} words, using Feature Agreement and Intersection over Union (IoU) as metrics. These scores are computed under both E-P and P-E prompting paradigms. To ensure fair comparison, agreement is measured only on instances where both agents assigned the same label to the input text.

Agreement scores are consistently higher in the P-E setting, indicating that post-prediction explanations align better with other models and human annotators. This highlights the impact of prompting order, motivating our exclusive focus on the P-E setting in subsequent analyses.

Model-to-model agreement exceeds 50\%, reflecting moderate consistency in identifying top-\textit{k} influential words. GPT-4o and DeepSeek-V3 exhibit the highest mutual agreement, suggesting similar interpretive strategies. In contrast, model-human agreement is consistently lower, highlighting fundamental differences in how humans and models prioritize influential words.

The consistently higher model-to-model agreement compared to model-human agreement suggests that LLMs may be converging on similar reasoning strategies, even when trained independently. This likely reflects their reliance on shared statistical regularities or keyword-based heuristics present in the training data, which makes their explanations mutually consistent. Humans, by contrast, draw on semantic understanding, pragmatic cues, and cultural context when identifying influential words, leading to explanations that often diverge from model-generated rationales. This divergence underscores a central challenge in explainability: achieving faithfulness requires more than internal model alignment; it demands bridging the gap between statistical shortcuts and the richer, context-driven reasoning used by humans. Notably, this finding reinforces our earlier hypothesis in Sec.~\ref{sec:EmotionClassificationPerformance}, where we argued that LLMs' superior benchmark performance reflects alignment with majoritarian statistical patterns rather than deeper semantic understanding. Taken together, these results highlight a systematic difference in how humans and LLMs reason about text. The challenge is especially pronounced in low-resource settings, where limited training data may aggravate reliance on shallow correlations, further widening the gap between model and human explanations.

\subsection{Qualitative Analysis} 
While quantitative metrics offer useful benchmarks, they cannot fully capture why model explanations diverge from human reasoning, even when both yield the same label.
Our qualitative examination of representative Persian samples reveals a consistent trend: humans emphasize affective or colloquial cues, whereas models attend to semantically central but emotionally neutral words.
This reflects a broader limitation of large language models, their reliance on lexical salience over socio-emotional understanding.
Such divergence is particularly pronounced in low-resource, colloquial-rich languages like Persian, underscoring the need for models that better integrate pragmatic and cultural cues.

\section{Conclusion}
We assessed the faithfulness of self-generated explanations from several auto-regressive LLMs using token-level log-probability analysis. Models were prompted to identify their top-\textit{k} influential words in emotion classification tasks on a Persian dataset, comparing the prompting paradigms E-P and P-E. Faithfulness metrics and agreement with human annotations were used for evaluation.

Temperature scaling improved model calibration, enhancing the alignment of model confidence with prediction reliability. GPT-family models outperformed native Persian annotators, showcasing their strength in low-resource languages, though no model achieved high faithfulness metrics. This suggests either the models' limitations in producing faithful explanations or the inadequacy of the chosen explanation strategy and evaluation metrics.

Models showed higher agreement with each other than with human annotations, indicating a divergence in reasoning. The P-E paradigm yielded more accurate classifications, faithful explanations, and closer alignment with human annotations, suggesting that post-hoc rationalization improves explanation quality.

These findings highlight the importance of model architecture and prompting strategy in generating faithful self-explanations. Future work should explore alternative explanation strategies and refine faithfulness metrics, as well as expand research to diverse datasets, languages, and tasks for more robust evaluations.

\section*{Limitations}
Our study has several limitations. First, we could not evaluate locally hosted models due to the high computational and memory demands of large-scale white-box models. Additionally, few models offer token-level log-probabilities through official APIs, limiting comparative analysis.

Second, our prompting paradigm introduces limitations. Although we used consistent prompts across models, LLMs' sensitivity to phrasing means different formulations could yield different predictions or explanations. Their inherent stochasticity may also lead to slight variations across runs, even under identical settings. While we carefully designed the prompts, we cannot guarantee perfect alignment with model interpretation.

Finally, the metrics employed are proxies and may not fully capture alignment with the models' internal reasoning. Therefore, our findings do not suggest that LLMs are incapable of faithful self-explanation, but rather that, under the specific strategies and metrics used in this study, they often fall short.

\section{Bibliographical References}
\bibliographystyle{lrec2026-natbib}
\bibliography{custom}

\appendix
\section{Human Annotation Protocol and Participant Details}\label{app:human}

\begin{table}[t]
    \centering
    \small
    \renewcommand{\arraystretch}{1.4}
    \setlength{\tabcolsep}{3.5pt}
    \begin{tabular}{llcc}
        \toprule
        \textbf{Model} & \textbf{Input Format} & \multicolumn{2}{c}{\textbf{Average Reduction}} \\
        \cmidrule(lr){3-4}
        & & \textbf{P-E} & \textbf{E-P} \\
        \midrule
                        & Full-text & 16.60 & 12.17 \\
        \textbf{GPT-4o} & Top-\textit{k} only & 8.77 & 7.07 \\
                        & Top-\textit{k} removed & 12.32 & 9.85 \\
        \midrule
                          & Full-text & 39.03 & 20.65 \\
        \textbf{GPT-4-turbo} & Top-\textit{k} only & 17.85 & 10.95 \\
                          & Top-\textit{k} removed & 9.67 & 5.03 \\
        \midrule
                       & Full-text & 36.82 & 35.22 \\
        {\parbox{2cm}{\textbf{Llama3.3\\70B-instruct}}} & Top-\textit{k} only & 18.71 & 14.99 \\
                       & Top-\textit{k} removed & 17.71 & 22.71 \\
        \midrule
                       & Full-text & 33.16 & 25.38 \\
        \textbf{DeepSeek-V3} & Top-\textit{k} only & 21.39 & 24.89 \\
                       & Top-\textit{k} removed & 18.48 & 14.34 \\
        \bottomrule
    \end{tabular}
    \captionsetup{width=\linewidth}
    \caption{Average model confidence score reduction of before and after temperature scaling on the evaluation set. Results are reported across different input formats (full-text, top-\textit{k} only, and top-\textit{k} removed) and prompting paradigms (P-E and E-P). Larger reductions reflect stronger correction of overconfident predictions. The results are reported as percentages.}
    \label{tab:confidencechangesummary}
\end{table}

\begin{table*}[h!]
    \centering
    \small
    \renewcommand{\arraystretch}{1.2}
    \setlength{\tabcolsep}{4.7pt}
    \begin{tabular}{lcccccc}
        \toprule
        \textbf{Model} & \textbf{Comp~(↑)} & \textbf{Suff~(↓)} & \textbf{DF\textsubscript{TopKRemoved}~(↑)} & \textbf{DF\textsubscript{TopKOnly}~(↓)} \\
        \midrule
        \multicolumn{5}{c}{\textbf{Predict-then-Explain (P-E)}} \\
        \midrule
        \textbf{GPT-4o} & 31.7 & \textbf{15.91} & 38.00 & \textbf{22.00} \\
        \textbf{GPT-4-turbo} & \textbf{47.11} & 19.60 & \textbf{53.33} & 24.33 \\
        \textbf{Llama3.3-70B-instruct} & 27.08 & 25.95 & 30.33 & 29.33 \\
        \textbf{DeepSeek-V3} & 33.48 & 24.52 & 39.67 & 29.00 \\
        \midrule
        \multicolumn{5}{c}{\textbf{Explain-then-Predict (E-P)}} \\
        \midrule
        \textbf{GPT-4o} & 25.13 & 29.86 & 31.67 & 36.00 \\
        \textbf{GPT-4-turbo} & \textbf{39.98} & 34.26 & 43.00 & 37.67 \\
        \textbf{Llama3.3-70B-instruct} & 24.73 & 38.81 & 27.00 & 42.00 \\
        \textbf{DeepSeek-V3} & 39.82 & \textbf{22.57} & \textbf{48.33} & \textbf{30.00} \\
        \midrule
        \textbf{Human Annotators} & - & - & 53.33 & 39.00 \\
        \bottomrule
    \end{tabular}
    \captionsetup{width=\linewidth}
    \caption{Faithfulness comparison of underlying LLMs in P-E and E-P settings, before confidence calibration. All metrics are reported as percentages.}
    \label{tab:faithfulness_comparison_before_calibration}
\end{table*}

Human annotations were collected from a group of 25 native Persian speakers, all undergraduate students in computer science and computer engineering. Participants included both male and female students and were recruited from an academic environment on a voluntary basis. All participants were informed about the purpose of the study and how their annotations would be used.

All annotations were gathered using Google Forms. In the first stage, annotators were assigned unique subsets of full-text samples. For each instance, they classified the text into one of six emotion categories and selected the top-\textit{k} influential words guiding their decision, following translated versions of the same instructions provided to the LLMs.

In the second stage, annotators were presented with two modified versions of previously unseen samples: (1) a sparse-input version containing only the top-\textit{k} words, and (2) a masked-input version in which the top-\textit{k} words were replaced by a placeholder token. As in the first stage, they classified each input into one of the six emotion categories.

To maintain annotation quality and prevent familiarity bias, no annotator saw the same instance across both stages. All annotations were independently verified through cross-checking by fellow annotators to ensure consistency and correctness. All participants consented to contribute voluntarily, and their responses were anonymized.

\section{Calibration Impact on Confidence Scores}\label{app:confidencechange}
Table~\ref{tab:confidencechangesummary} summarizes the average reduction in confidence scores across input types and prompting paradigms. Reductions were most pronounced for full-text inputs, producing the largest confidence decreases.

\section{Faithfulness Evaluation Before Calibration}\label{app:faith}
Table~\ref{tab:faithfulness_comparison_before_calibration} presents the self-explanation faithfulness metrics for all models under both prompting paradigms, prior to confidence calibration. These results provide a baseline for understanding the raw behavior of each LLM before post-processing adjustments to confidence scores.

\section{Prompt Templates for API-Based LLM Interactions}\label{app:prompt}
This section outlines the standardized prompt templates used to interact with LLMs during evaluation. All interactions followed a structured role-based format involving \texttt{system}, \texttt{user}, and \texttt{assistant} messages.

Table~\ref{tab:predict_label_from_fulltext} shows the prompt used for classifying the full input text.
Table~\ref{tab:get_top_k} illustrates how we extracted the top-$k$ influential words (self-explanations), without requesting an emotion label.
Table~\ref{tab:predict_label_from_topk} presents the prompt used to classify based solely on the extracted influential words.
Table~\ref{tab:predict_label_from_topkremoved} contains the prompt used for classification after masking these words in the original input.

Table~\ref{tab:P-E_Full_Example} and~\ref{tab:E-P_Full_Example}  consolidates the full interaction flow under the Predict-then-Explain (P-E) and Explain-then-Predict (E-P) paradigm for a single evaluation sample, including the Persian input text, Chat GPT-4o model outputs, and word masking stages.

Note that in rare instances (fewer than 1\%) where the model output did not conform to the expected format (e.g., producing additional text), these samples were discarded to ensure the integrity and consistency of our reported metrics.

\section{Emotion Prediction Confusion Matrices}\label{app:confusion}

\begin{figure}
\centering
\begin{subfigure}{0.37\textwidth}
    \includegraphics[width=\textwidth]{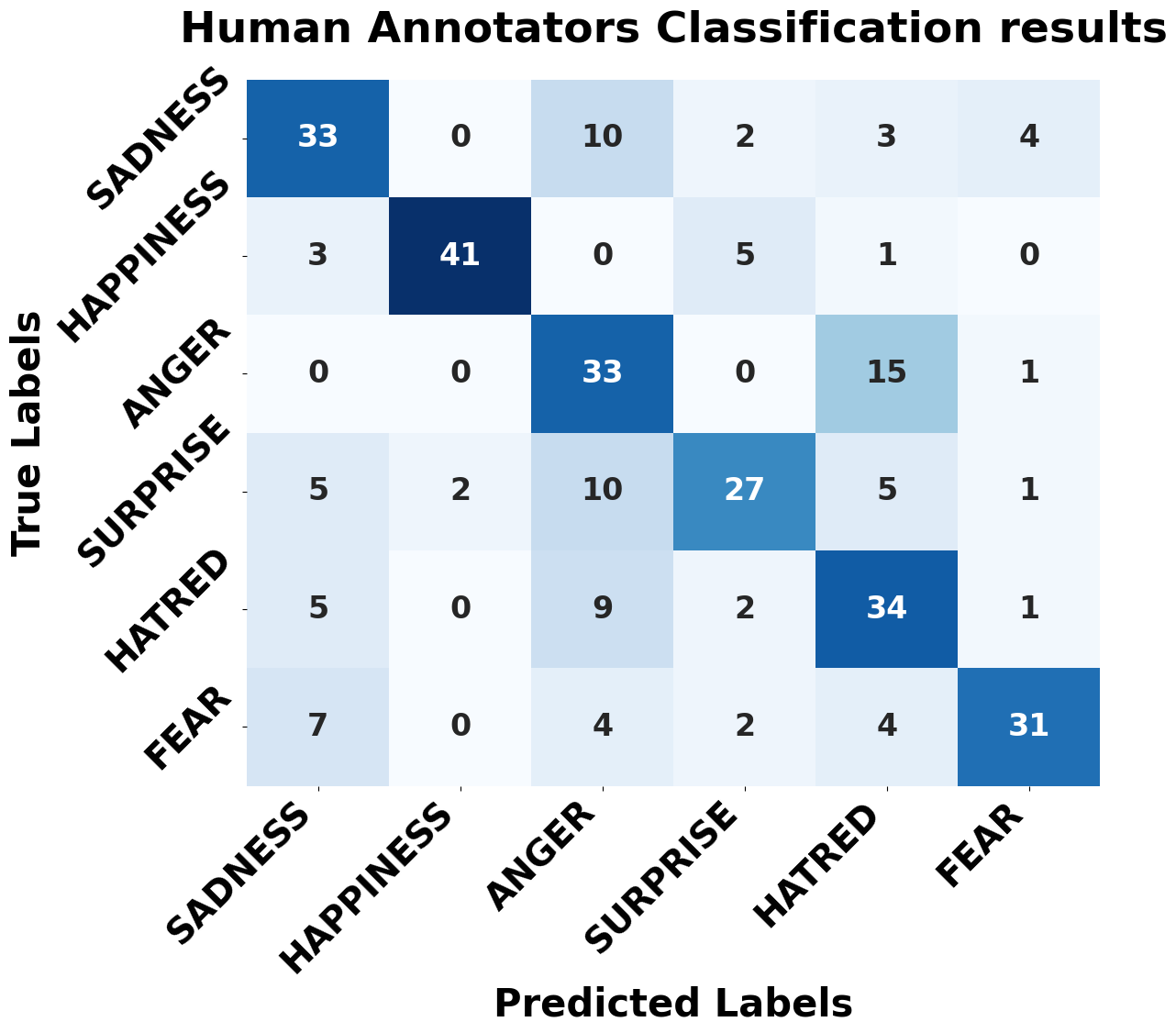}
\end{subfigure}
\caption{Confusion matrices of Human Annotator performance.}
\label{fig:CM-human}
\end{figure}

\begin{figure*}
    \centering    \includegraphics[width=\textwidth,height=0.9\textheight,keepaspectratio]{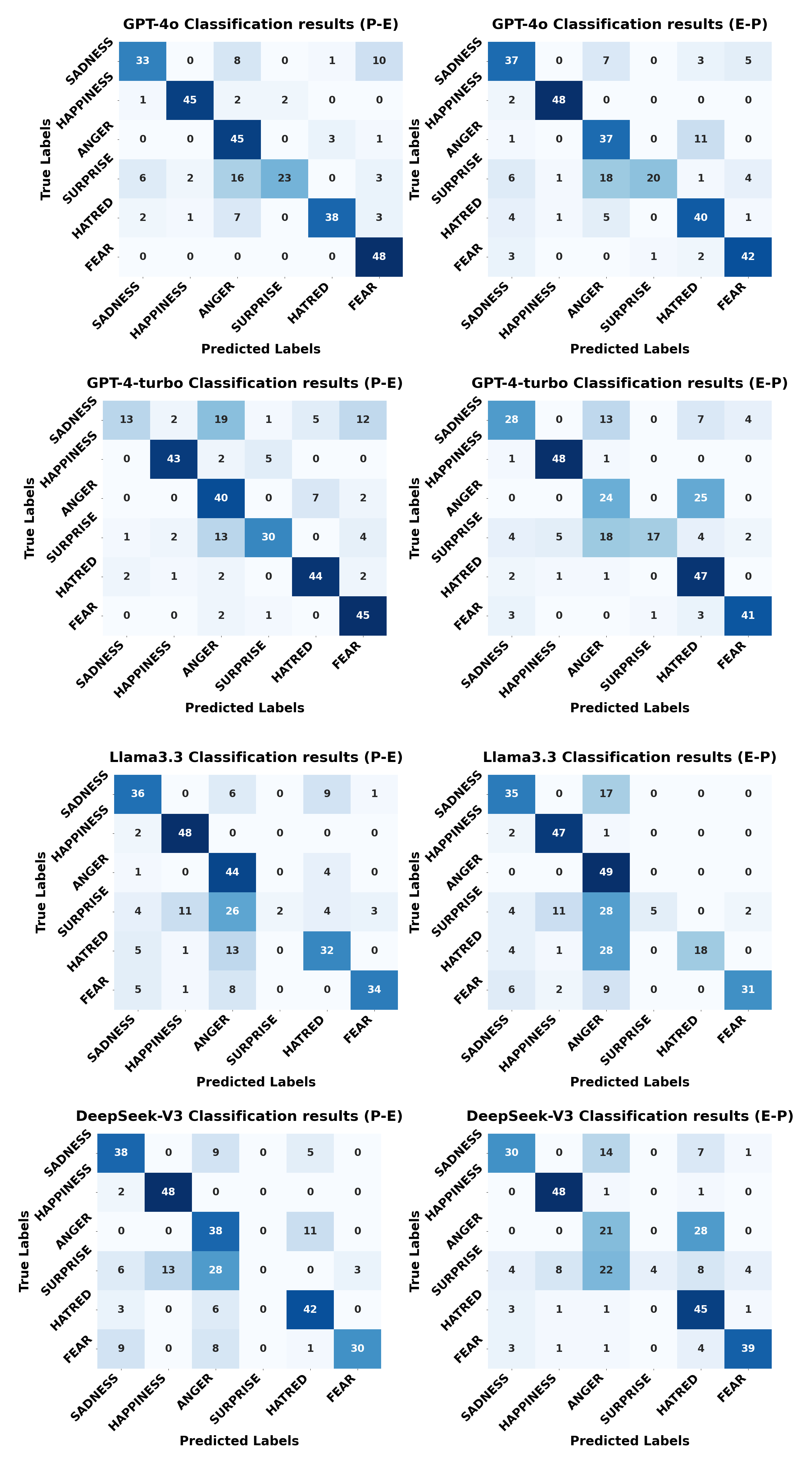}
    \captionsetup{width=\linewidth}
    \caption{Confusion matrices of different models in both P-E and E-P paradigms}
    \label{fig:CMs}
\end{figure*}

Figure~\ref{fig:CMs} displays the confusion matrices for all evaluated models under both P-E and E-P prompting conditions. Figure~\ref{fig:CM-human} shows the corresponding confusion matrix for human annotators.

\section{Reliability Diagrams}\label{app:reliable}

To visually assess the impact of temperature scaling on confidence calibration, we present reliability diagrams for both the calibration and evaluation sets. For visualization, we used the implementation provided by \citet{10.5555/3305381.3305518}, available at their official GitHub repository\footnote{\url{https://github.com/hollance/reliability-diagrams?tab=readme-ov-file}} with 20 bins for all models and prompting paradigms.

Prior to calibration, the models exhibit significant overconfidence. Most predictions have confidence scores close to 1, while actual accuracy lags behind, as shown by the noticeable gap between the red confidence bars and the black accuracy bars in the diagrams. This misalignment results in high \emph{Expected Calibration Error} (ECE) values.

\begin{table*}[]
    \small
    \renewcommand{\arraystretch}{1.5}
    \begin{tabular}{lp{0.9\linewidth}}
        \hline
        \rowcolor{gray!20} \textbf{Role} & \textbf{Content} \\
        \hline
        \rowcolor{blue!15} system & You are an emotion classifier. You must classify the emotion and output the top influential words in CSV format. For classifying, you are strictly required to output only one of the following English numbers: 0, 1, 2, 3, 4, or 5. No other output is acceptable. For top influential words, you can only output Persian words in the text. \\
        \hline
        \rowcolor{yellow!15} user & Classify the following text into one of the categories: 'Sadness':0, 'Happiness':1, 'Anger':2, 'Surprise':3, 'Hatred':4, 'Fear':5. For each class, output the mapped number. Only output an English number showing the class of the text. Make sure not to output any other character. Text: \\
    \hline
    \end{tabular}
    \captionsetup{width=\linewidth}
    \caption{The prompt used for the LLM for emotion classification of the original text}
    \label{tab:predict_label_from_fulltext}
\end{table*}

\begin{table*}[]
    \small
    \renewcommand{\arraystretch}{1.5}
    \begin{tabular}{lp{0.9\linewidth}}
        \hline
        \rowcolor{gray!20} \textbf{Role} & \textbf{Content} \\
        \hline
        \rowcolor{blue!15} system & You are an emotion classifier. You must classify the emotion and output the top influential words in CSV format. For classifying, you are strictly required to output only one of the following English numbers: 0, 1, 2, 3, 4, or 5. No other output is acceptable. For top influential words, you can only output Persian words in the text. \\
        \hline
        \rowcolor{yellow!15} user & List the top {K} most influential words that contributed to this classification in CSV format (in a single line). I don't want you to classify in this stage. Only give me top {K} words. Make sure to provide only {K} Persian words which they exist in the original text and don't output any other token. Here is an example output: word1,word2,word3,word4,word5 Text: \\
        \hline
    \end{tabular}
    \captionsetup{width=\linewidth}
    \caption{The prompt used for the LLM to get the most influential words from the original text as the self-explanation}
    \label{tab:get_top_k}
\end{table*}

\begin{table*}[]
    \small
    \renewcommand{\arraystretch}{1.5}
    \begin{tabular}{lp{0.9\linewidth}}
        \hline
        \rowcolor{gray!20} \textbf{Role} & \textbf{Content} \\
        \hline
        \rowcolor{blue!15} system & You are an emotion classifier. You are provided with some influential words that have been extracted from the text. You must classify the emotion based only on these words. For classifying, you are strictly required to output only one of the following English numbers: 0, 1, 2, 3, 4, or 5. No other output is acceptable. \\
        \hline
        \rowcolor{yellow!15} user & Classify the following text into one of the categories: 'Sadness':0, 'Happiness':1, 'Anger':2, 'Surprise':3, 'Hatred':4, 'Fear':5. For each class output the mapped number. Only output an English number showing the class of the text. Make sure not to output any other character. Text: \\ 
        \hline
    \end{tabular}
    \captionsetup{width=\linewidth}
    \caption{The prompt used for the LLM for emotion classification of the top-\textit{k} most influential words alone}
    \label{tab:predict_label_from_topk}
\end{table*}

\begin{table*}[]
    \small
    \renewcommand{\arraystretch}{1.5}
    \begin{tabular}{lp{0.9\linewidth}}
        \hline
        \rowcolor{gray!20} \textbf{Role} & \textbf{Content} \\
        \hline
        \rowcolor{blue!15} system & You are an emotion classifier. In the text, some influential words have been replaced with the placeholder [removed]. You must classify the emotion based on the text, considering these [removed] words as part of the context. For classifying, you are strictly required to output only one of the following English numbers: 0, 1, 2, 3, 4, or 5. No other output is acceptable. \\
        \hline
        \rowcolor{yellow!15} user & Classify the following text into one of the categories: 'Sadness':0, 'Happiness':1, 'Anger':2, 'Surprise':3, 'Hatred':4, 'Fear':5. For each class output the mapped number. Only output an English number showing the class of the text. Make sure not to output any other character. Text: \\ 
        \hline
    \end{tabular}
    \captionsetup{width=\linewidth}
    \caption{The prompt used for the LLM for emotion classification after replacing the top-\textit{k} most influential words in the original text with placeholders} 
    \label{tab:predict_label_from_topkremoved}
\end{table*}

\renewcommand{\FR}[1]{{\footnotesize\textRL{#1}}}
\begin{table*}[]
\small
\renewcommand{\arraystretch}{1.5}
\begin{threeparttable}
\begin{tabular}{lp{0.875\linewidth}}
    \hline
    \rowcolor{gray!20} \textbf{Role} & \textbf{Content} \\
    \hline
    \rowcolor{blue!15} system & You are an emotion classifier. You must classify the emotion and output the top influential words in CSV format. For classifying, you are strictly required to output only one of the following English numbers: 0, 1, 2, 3, 4, or 5. No other output is acceptable. For top influential words, you can only output Persian words in the text. \\
    \rowcolor{yellow!15} user & Classify the following text into one of the categories: 'Sadness':0, 'Happiness':1, 'Anger':2, 'Surprise':3, 'Hatred':4, 'Fear':5. For each class, output the mapped number. Only output an English number showing the class of the text. Make sure not to output any other character. \\
    \rowcolor{yellow!15} & Text: {\FR{\small\footnotesizeدلم اون حالیو میخواد که اونقدر فکرم رها باشه اونقدر سرگرم خوندن شاهکارای ادبی دنیا باشم ساعت ۳ یهو ب پنجره نگاه کنم ببینم برف میاد و ده ها برابر حالم خوب شه و با اشتیاق ب خوندن ادامه بدم}}\tnote{a} \\
    \rowcolor{green!15} assistant & 1 \\
    \rowcolor{yellow!15} user & Then, list the top K most influential words that contributed to this classification in CSV format (in a single line). Make sure to provide only K Persian words that exist in the original text and don't output any other token. \\
    \rowcolor{yellow!15} & Text: {\FR{\small\footnotesizeدلم اون حالیو میخواد که اونقدر فکرم رها باشه اونقدر سرگرم خوندن شاهکارای ادبی دنیا باشم ساعت ۳ یهو ب پنجره نگاه کنم ببینم برف میاد و ده ها برابر حالم خوب شه و با اشتیاق ب خوندن ادامه بدم}}\tnote{a} \\
    \rowcolor{green!15} assistant & {\FR{\small\footnotesizeادامه}}, {\FR{\small\footnotesizeاشتیاق}}, {\FR{\small\footnotesizeخوب}}, {\FR{\small\footnotesizeسرگرم}}, {\FR{\small\footnotesizeدلم}}\tnote{b} \\
    \hline
    \rowcolor{blue!15} system & You are an emotion classifier. You are provided with some influential words that have been extracted from the text. You must classify the emotion based only on these words. For classifying, you are strictly required to output only one of the following English numbers: 0, 1, 2, 3, 4, or 5. No other output is acceptable. \\
    \rowcolor{yellow!15} user & Classify the following text into one of the categories: 'Sadness':0, 'Happiness':1, 'Anger':2, 'Surprise':3, 'Hatred':4, 'Fear':5. For each class output the mapped number. Only output an English number showing the class of the text. Make sure not to output any other character. Text: \\ 
    \rowcolor{yellow!15} & Text: {\FR{\small\footnotesizeادامه}}, {\FR{\small\footnotesizeاشتیاق}}, {\FR{\small\footnotesizeخوب}}, {\FR{\small\footnotesizeسرگرم}}, {\FR{دلم\small\footnotesize}}\tnote{b}  \\
    \rowcolor{green!15} assistant & 1 \\
    \hline
    \rowcolor{blue!15} system & You are an emotion classifier. In the text, some influential words have been replaced with the placeholder [\FR{\small\footnotesizeحذف شده}]\tnote{c}. You must classify the emotion based on the text, considering these [\FR{\small\footnotesizeحذف شده}]\tnote{c} words as part of the context. For classifying, you are strictly required to output only one of the following English numbers: 0, 1, 2, 3, 4, or 5. No other output is acceptable. \\
    \rowcolor{yellow!15} user & Classify the following text into one of the categories: 'Sadness':0, 'Happiness':1, 'Anger':2, 'Surprise':3, 'Hatred':4, 'Fear':5. For each class output the mapped number. Only output an English number showing the class of the text. Make sure not to output any other character. \\ 
    \rowcolor{yellow!15} & Text:{\FR{\small\footnotesize[حذف شده] اون حالیو میخواد که اونقدر فکرم رها باشه اونقدر [حذف شده] خوندن شاهکارای ادبی دنیا باشم ساعت ۳ یهو ب پنجره نگاه کنم ببینم برف میاد و ده ها برابر حالم [حذف شده] شه و با [حذف شده] ب خوندن [حذف شده] بدم}} \tnote{d} \\
    \rowcolor{green!15} assistant & 0 \\
    \hline
\end{tabular}
\begin{tablenotes}
    \small\footnotesize
    \begin{minipage}{0.98\linewidth}
    \item[a] \textbf{Translation:} "My heart wants that feeling where my mind is so free, I'm so engrossed in reading the literary masterpieces of the world, and suddenly at 3 o'clock I look at the window and see snow falling, and my mood gets many times better, and I continue reading with enthusiasm." %\\
    \textbf{Transliteration:} delam un hâli-o mikhâd ke onghadr fekram rahâ bâshe, onghadr sargarm-e khundan-e shâhkârâ-ye adabi-ye donyâ bâsham, sâ'at-e 3 yeho be panjare negâh konam, bebinam barf miyâd, va dah-hâ barâbar hâlam khub she, va bâ eshtiâq be khundan edâme bedam.
    \item[b] \textbf{Translation:} edâme 'continuation', eshtiâq 'enthusiasm', khub 'good', sargarm 'engaged', delam 'my heart'.
    \item[c] \textbf{Translation:} "Removed" \textbf{Explanation:} hazf shode.
    \item[d] \textbf{Translation:} "[removed] wants that feeling where my mind is so free, I'm so [removed] in reading the literary masterpieces of the world, and suddenly at 3 o'clock I look at the window and see snow falling, and my mood gets many times [removed], and I continue reading with [removed]." %\\
    \textbf{Transliteration:} [hazf shode] un hâli-o mikhâd ke onghadr fekram rahâ bâshe, onghadr [hazf shode] khundan-e shâhkârâ-ye adabi-ye donyâ bâsham, sâ'at-e 3 yeho be panjare negâh konam, bebinam barf miyâd, va dah-hâ barâbar hâlam [hazf shode] she, va bâ [hazf shode] be khundan [hazf shode] bedam. \\
    \end{minipage}
\end{tablenotes}
\end{threeparttable}
\captionsetup{width=\linewidth}
\caption{ This table illustrates the complete interaction protocol between system, user, and assistant roles for a single data sample during evaluation. It covers the Predict-then-Explain (P-E) paradigm across four stages: (1) standard classification, (2) influential word extraction, and (3) classification using only influential words (4) classification using modified input with influential words removed.}
\label{tab:P-E_Full_Example}
\end{table*}

\renewcommand{\FR}[1]{{\footnotesize\textRL{#1}}}
\begin{table*}[]
\small
\renewcommand{\arraystretch}{1.5}
\begin{threeparttable}
\begin{tabular}{lp{0.875\linewidth}}
    \hline
    \rowcolor{gray!20} \textbf{Role} & \textbf{Content} \\
    \hline
    \rowcolor{blue!15} system & You are an emotion classifier. You must classify the emotion and output the top influential words in CSV format. For classifying, you are strictly required to output only one of the following English numbers: 0, 1, 2, 3, 4, or 5. No other output is acceptable. For top influential words, you can only output Persian words in the text. \\
    \rowcolor{yellow!15} user & List the top K most influential words that contributed to this classification in CSV format (in a single line). Make sure to provide only K Persian words that exist in the original text and don't output any other token. \\
    \rowcolor{yellow!15} & Text: {\FR{\small\footnotesizeدلم اون حالیو میخواد که اونقدر فکرم رها باشه اونقدر سرگرم خوندن شاهکارای ادبی دنیا باشم ساعت ۳ یهو ب پنجره نگاه کنم ببینم برف میاد و ده ها برابر حالم خوب شه و با اشتیاق ب خوندن ادامه بدم}}\tnote{a} \\
    \rowcolor{green!15} assistant &  {\FR{\small\footnotesizeادامه}}, {\FR{\small\footnotesizeاشتیاق}}, {\FR{\small\footnotesizeخوب}}, {\FR{\small\footnotesizeسرگرم}}, {\FR{\small\footnotesizeدلم}}\tnote{b} \\
    \rowcolor{yellow!15} user & Then, Classify the following text into one of the categories: 'Sadness':0, 'Happiness':1, 'Anger':2, 'Surprise':3, 'Hatred':4, 'Fear':5. For each class, output the mapped number. Only output an English number showing the class of the text. Make sure not to output any other character. \\
    \rowcolor{yellow!15} & Text: {\FR{\small\footnotesizeدلم اون حالیو میخواد که اونقدر فکرم رها باشه اونقدر سرگرم خوندن شاهکارای ادبی دنیا باشم ساعت ۳ یهو ب پنجره نگاه کنم ببینم برف میاد و ده ها برابر حالم خوب شه و با اشتیاق ب خوندن ادامه بدم}}\tnote{a} \\
    \rowcolor{green!15} assistant & 1 \\
    \hline
    \rowcolor{blue!15} system & You are an emotion classifier. You are provided with some influential words that have been extracted from the text. You must classify the emotion based only on these words. For classifying, you are strictly required to output only one of the following English numbers: 0, 1, 2, 3, 4, or 5. No other output is acceptable. \\
    \rowcolor{yellow!15} user & Classify the following text into one of the categories: 'Sadness':0, 'Happiness':1, 'Anger':2, 'Surprise':3, 'Hatred':4, 'Fear':5. For each class output the mapped number. Only output an English number showing the class of the text. Make sure not to output any other character. Text: \\ 
    \rowcolor{yellow!15} & Text: {\FR{\small\footnotesizeادامه}}, {\FR{\small\footnotesizeاشتیاق}}, {\FR{\small\footnotesizeخوب}}, {\FR{\small\footnotesizeسرگرم}}, {\FR{دلم\small\footnotesize}}\tnote{b}  \\
    \rowcolor{green!15} assistant & 1 \\
    \hline
    \rowcolor{blue!15} system & You are an emotion classifier. In the text, some influential words have been replaced with the placeholder [\FR{\small\footnotesizeحذف شده}]\tnote{c}. You must classify the emotion based on the text, considering these [\FR{\small\footnotesizeحذف شده}]\tnote{c} words as part of the context. For classifying, you are strictly required to output only one of the following English numbers: 0, 1, 2, 3, 4, or 5. No other output is acceptable. \\
    \rowcolor{yellow!15} user & Classify the following text into one of the categories: 'Sadness':0, 'Happiness':1, 'Anger':2, 'Surprise':3, 'Hatred':4, 'Fear':5. For each class output the mapped number. Only output an English number showing the class of the text. Make sure not to output any other character. \\ 
    \rowcolor{yellow!15} & Text:{\FR{\small\footnotesize[حذف شده] اون حالیو میخواد که اونقدر فکرم رها باشه اونقدر [حذف شده] خوندن شاهکارای ادبی دنیا باشم ساعت ۳ یهو ب پنجره نگاه کنم ببینم برف میاد و ده ها برابر حالم [حذف شده] شه و با [حذف شده] ب خوندن [حذف شده] بدم}} \tnote{d} \\
    \rowcolor{green!15} assistant & 0 \\
    \hline
\end{tabular}
\begin{tablenotes}
    \small\footnotesize
    \begin{minipage}{0.98\linewidth}
    \item[a] \textbf{Translation:} "My heart wants that feeling where my mind is so free, I'm so engrossed in reading the literary masterpieces of the world, and suddenly at 3 o'clock I look at the window and see snow falling, and my mood gets many times better, and I continue reading with enthusiasm." %\\
    \textbf{Transliteration:} delam un hâli-o mikhâd ke onghadr fekram rahâ bâshe, onghadr sargarm-e khundan-e shâhkârâ-ye adabi-ye donyâ bâsham, sâ'at-e 3 yeho be panjare negâh konam, bebinam barf miyâd, va dah-hâ barâbar hâlam khub she, va bâ eshtiâq be khundan edâme bedam.
    \item[b] \textbf{Translation:} edâme 'continuation', eshtiâq 'enthusiasm', khub 'good', sargarm 'engaged', delam 'my heart'.
    \item[c] \textbf{Translation:} "Removed" \textbf{Explanation:} hazf shode.
    \item[d] \textbf{Translation:} "[removed] wants that feeling where my mind is so free, I'm so [removed] in reading the literary masterpieces of the world, and suddenly at 3 o'clock I look at the window and see snow falling, and my mood gets many times [removed], and I continue reading with [removed]." %\\
    \textbf{Transliteration:} [hazf shode] un hâli-o mikhâd ke onghadr fekram rahâ bâshe, onghadr [hazf shode] khundan-e shâhkârâ-ye adabi-ye donyâ bâsham, sâ'at-e 3 yeho be panjare negâh konam, bebinam barf miyâd, va dah-hâ barâbar hâlam [hazf shode] she, va bâ [hazf shode] be khundan [hazf shode] bedam. \\
    \end{minipage}
\end{tablenotes}
\end{threeparttable}
\captionsetup{width=\linewidth}
\caption{ This table illustrates the complete interaction protocol between system, user, and assistant roles for a single data sample during evaluation. It covers the Explain-then-Predict (E-P) paradigm across four stages: (1) influential word extraction, (2) standard classification, and (3) classification using only influential words (4) classification using modified input with influential words removed.}
\label{tab:E-P_Full_Example}
\end{table*}

After applying temperature scaling, this discrepancy is substantially reduced. The confidence scores become more evenly distributed across the [0, 1] interval, and the gap between predicted confidence and actual accuracy narrows considerably. This leads to a marked reduction in ECE and a better-calibrated model, where confidence scores are more trustworthy indicators of predictive accuracy.

This improvement generalizes to the evaluation set as well. Using the \emph{temperature} parameters optimized for each model on the calibration set, we observe a similar alignment between the confidence of each model and accuracy in evaluation, confirming the effectiveness of temperature scaling across datasets.

%%..%%..%%..%%..%%..%%..%%..%%..%%
\begin{figure*}
    \includegraphics[width=\textwidth, height=0.35\textwidth]{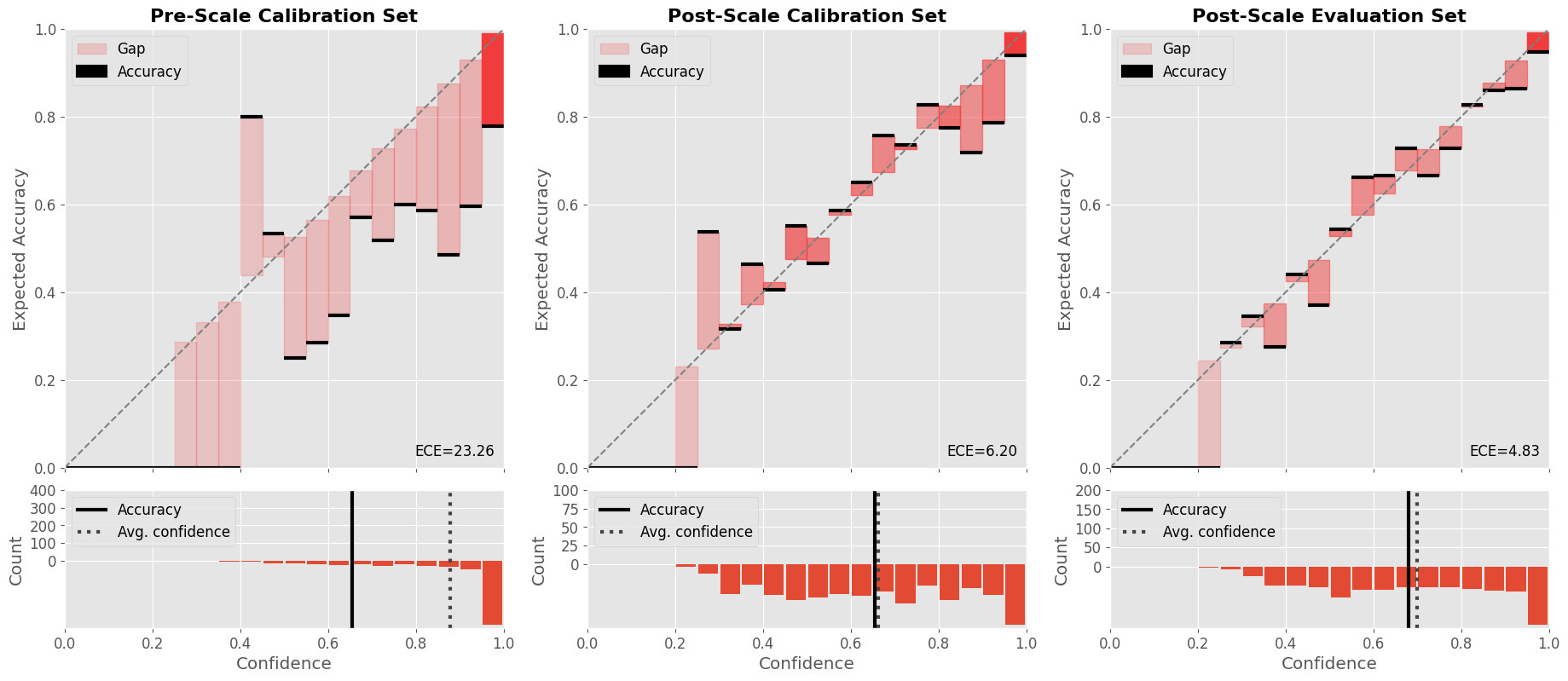}
    \captionsetup{width=\linewidth}
    \caption{Reliability diagrams of the Chat GPT-4o model under P-E paradigm.}
    \label{fig:4o_PE_REL}
\end{figure*}
\begin{figure*}
    \includegraphics[width=\textwidth, height=0.35\textwidth]{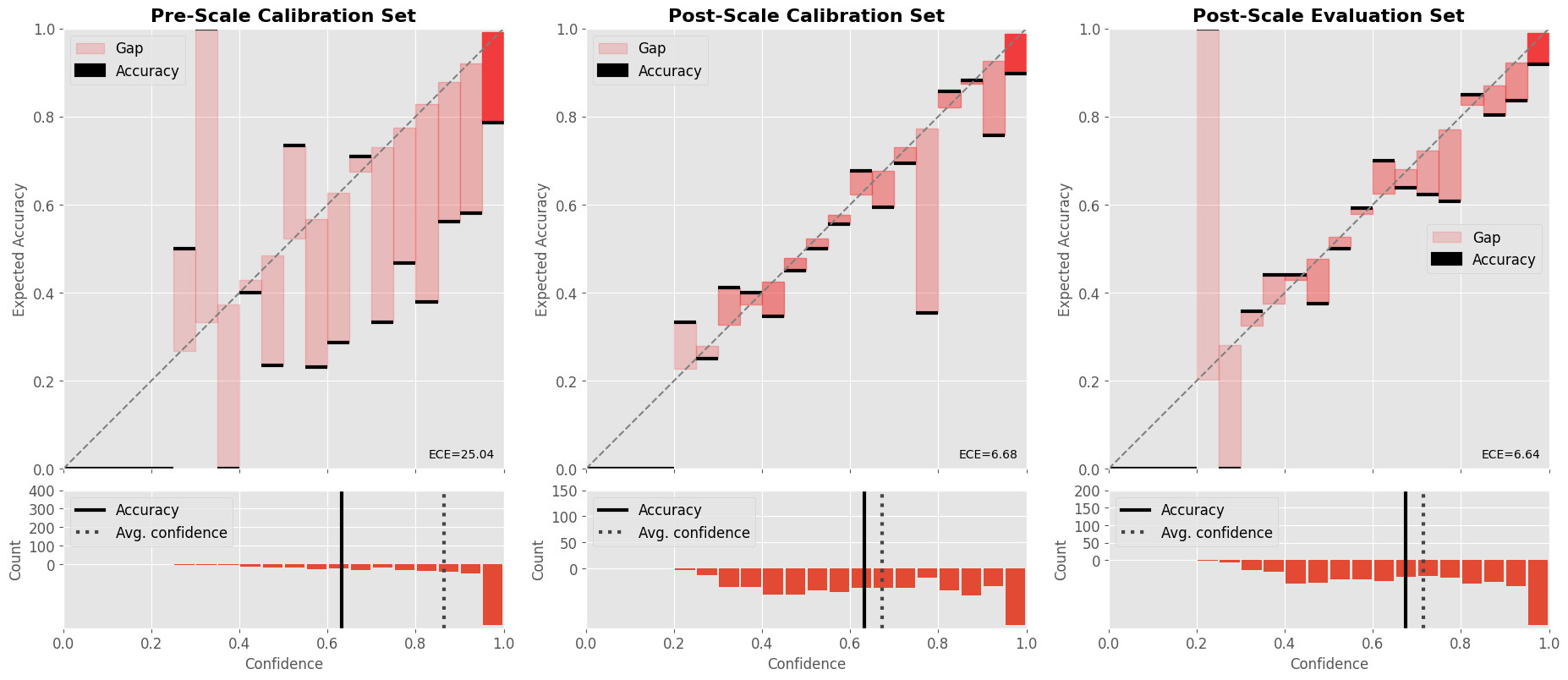}
    \captionsetup{width=\linewidth}
    \caption{Reliability diagrams of the Chat GPT-4o model under E-P paradigm.}
    \label{fig:4o_EP_REL}
\end{figure*}
%%..%%..%%..%%..%%..%%..%%..%%..%%
\begin{figure*}
    \includegraphics[width=\textwidth, height=0.35\textwidth]{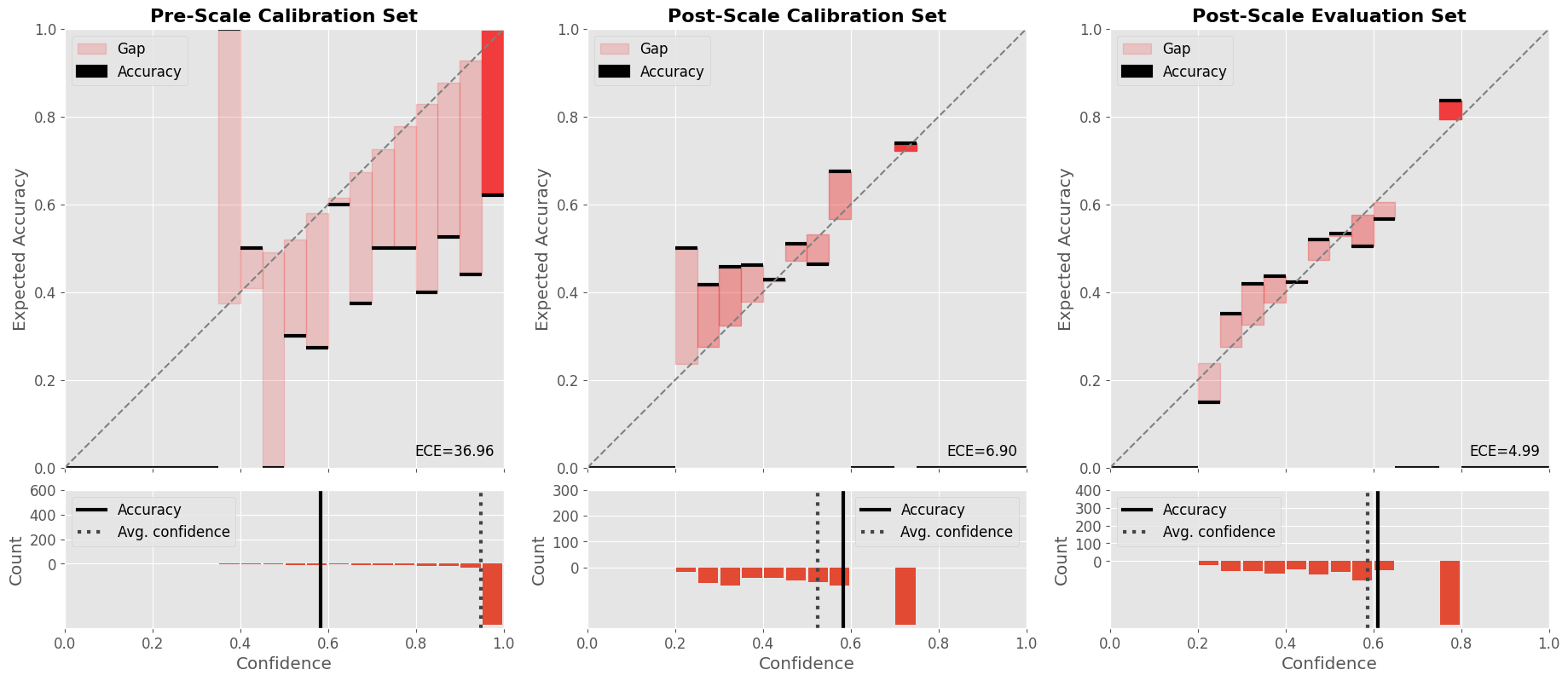}
    \captionsetup{width=\linewidth}
    \caption{Reliability diagrams of the Chat GPT-4-turbo model under P-E paradigm.}
    \label{fig:4_PE_REL}
\end{figure*}
\begin{figure*}
    \includegraphics[width=\textwidth, height=0.35\textwidth]{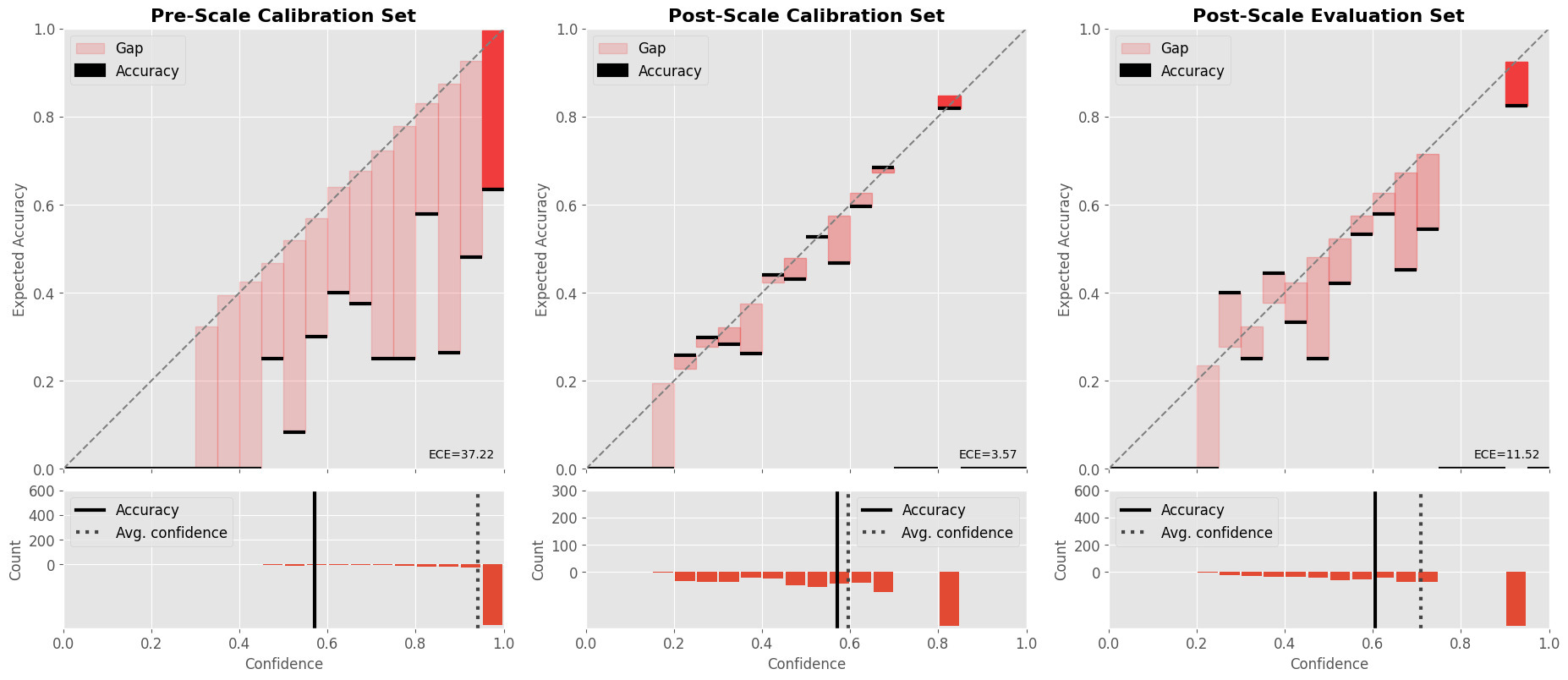}
    \captionsetup{width=\linewidth}
    \caption{Reliability diagrams of the Chat GPT-4-turbo model under E-P paradigm.}
    \label{fig:4_EP_REL}
\end{figure*}
%%..%%..%%..%%..%%..%%..%%..%%..%%
\begin{figure*}
    \includegraphics[width=\textwidth, height=0.35\textwidth]{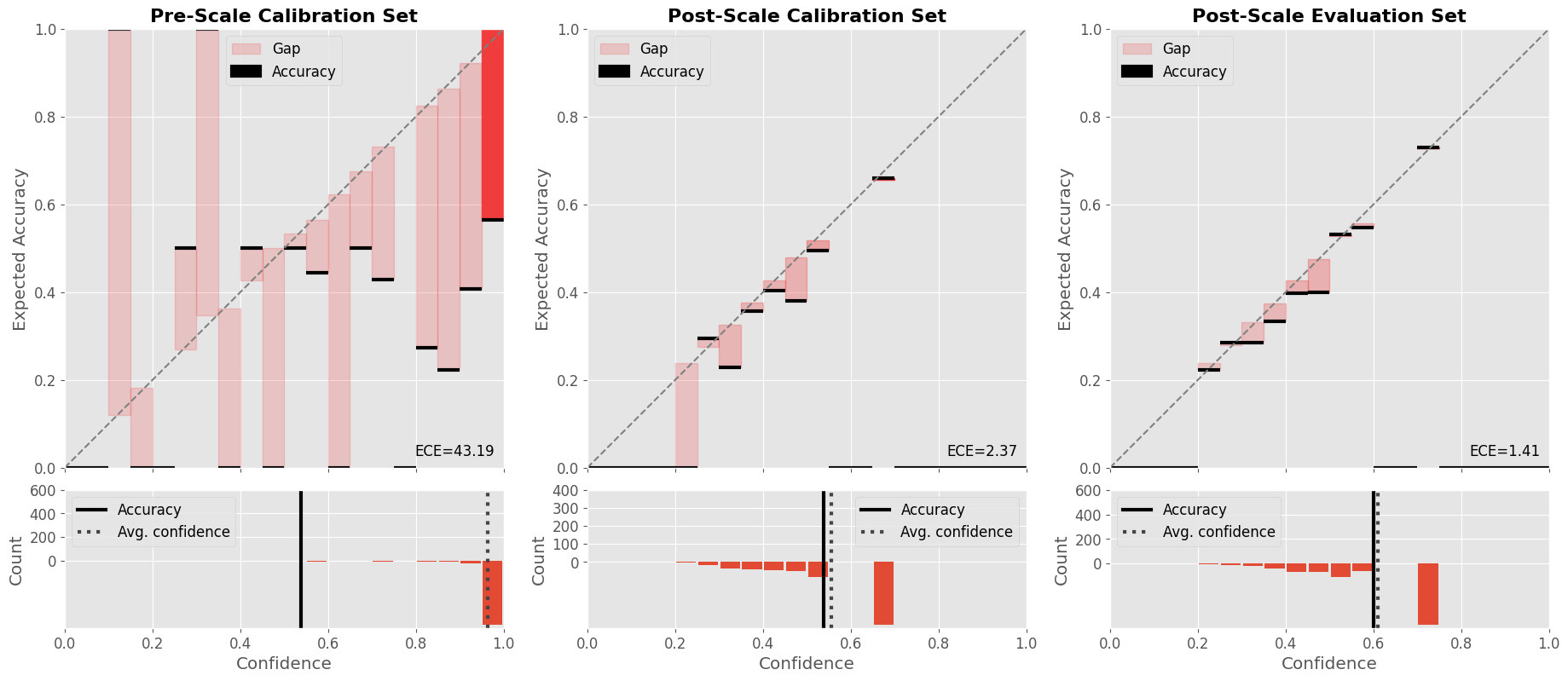}
    \captionsetup{width=\linewidth}
    \caption{Reliability diagrams of the Chat Llama3.3-70B-instruct model under P-E paradigm.}
    \label{fig:LL_PE_REL}
\end{figure*}
\begin{figure*}
    \includegraphics[width=\textwidth, height=0.35\textwidth]{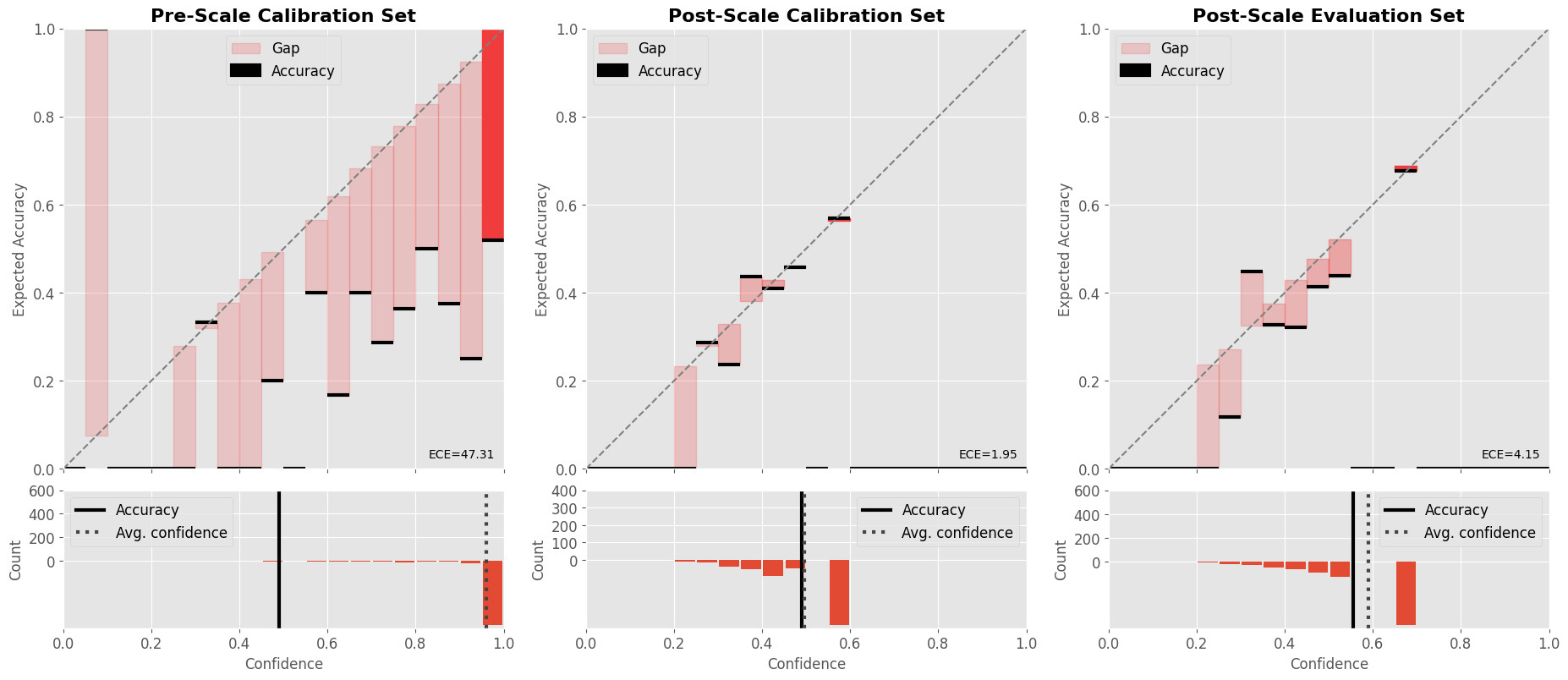}
    \captionsetup{width=\linewidth}
    \caption{Reliability diagrams of the Chat Llama3.3-70B-instruct model under E-P paradigm.}
    \label{fig:LL_EP_REL}
\end{figure*}
%%..%%..%%..%%..%%..%%..%%..%%..%%
\begin{figure*}
    \includegraphics[width=\textwidth, height=0.35\textwidth]{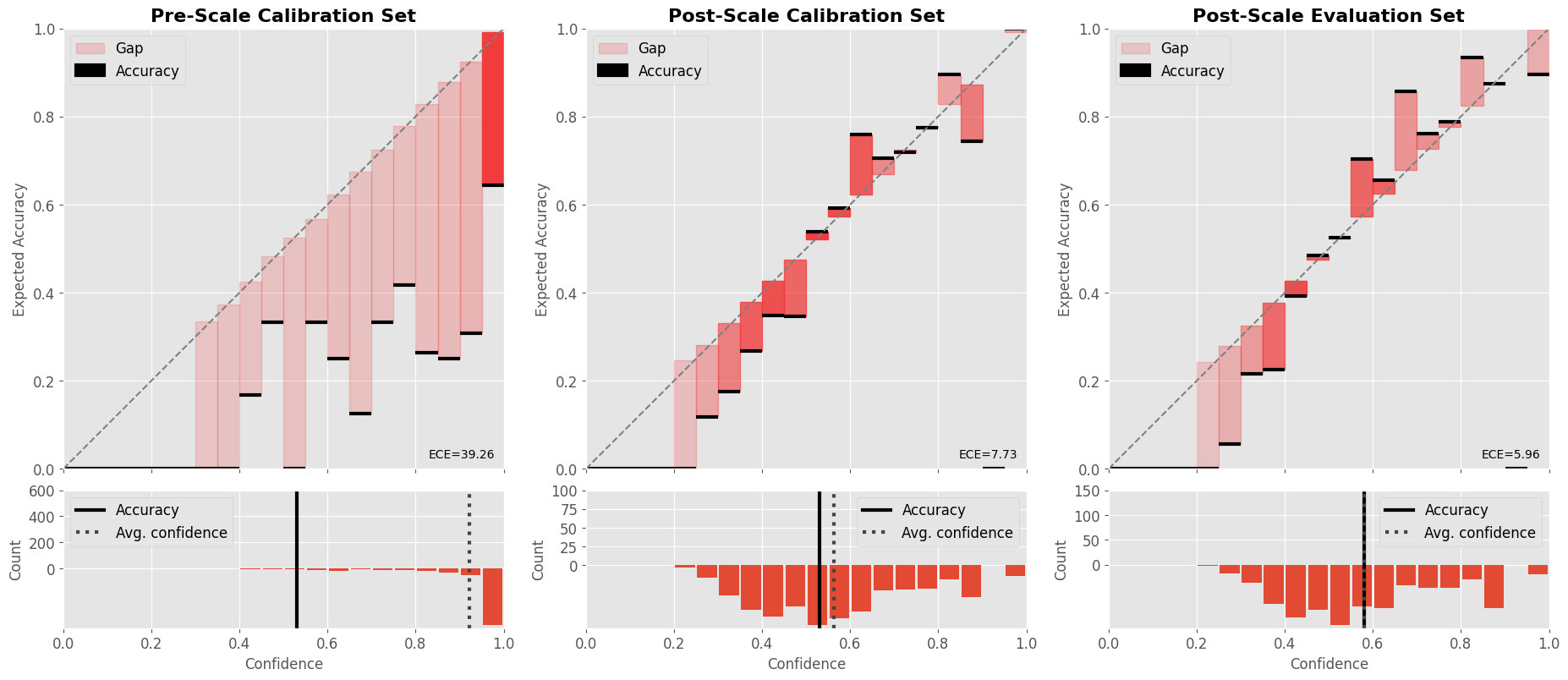}
    \captionsetup{width=\linewidth}
    \caption{Reliability diagrams of the DeepSeek-V3 model under P-E paradigm.}
    \label{fig:DS_PE_REL}
\end{figure*}
\begin{figure*}
    \includegraphics[width=\textwidth, height=0.35\textwidth]{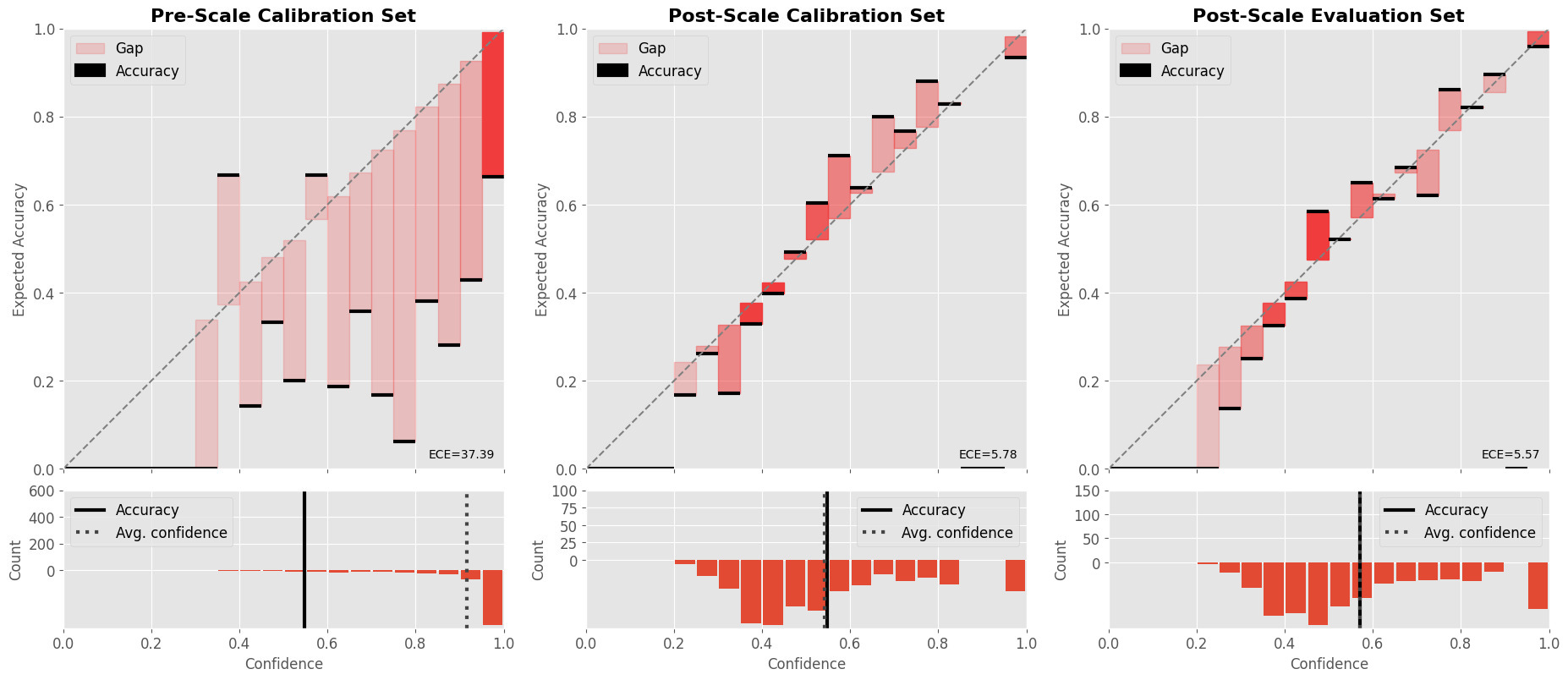}
    \captionsetup{width=\linewidth}
    \caption{Reliability diagrams of the DeepSeek-V3 model under E-P paradigm.}
    \label{fig:DS_EP_REL}
\end{figure*}
%%..%%..%%..%%..%%..%%..%%..%%..%%

\section{Qualitative Examples of Human-Model Divergence}\label{app:qualitative-analysis}
As shown in Table~\ref{tab:faithfulness_comparison_before_calibration}, GPT-family models (GPT-4o and GPT-4-turbo) consistently outperform Llama3.3-70B-instruct and DeepSeek-V3 in faithfulness metrics. For this reason, the qualitative examples presented here focus exclusively on GPT models in comparison with human annotations. This enables us to better isolate the key differences between state-of-the-art model explanations and human reasoning.

Across examples, we observe that models often prioritize topical or semantically central terms, while human annotators highlight colloquial, affective, and contextually charged words. This divergence helps explain why model explanations, though computationally faithful, may still fail to align with human reasoning.

Table~\ref{tab:qualitative-gpt4t} presents representative samples comparing GPT-4-turbo and human annotations under both Predict-then-Explain (P-E) and Explain-then-Predict (E-P) settings. We observe that GPT-4-turbo frequently highlights structurally or thematically relevant words (e.g., \FR{کتابخون} : "well-read", \FR{مامانت} : "your mom"), whereas human annotators consistently point to affective slang or evaluative markers (e.g., \FR{خر} : "tupid", \FR{گند} : "mess") as decisive signals of emotion. Even in positive cases, such as texts labeled Happiness, humans emphasize intensifiers like \FR{بیش باد} ("may it increase") that convey strong cultural and emotional resonance, while the model focuses on neutral topical references such as named entities.

Table~\ref{tab:qualitative-gpt4o} provides analogous examples for GPT-4o. Here we see a similar trend: the model tends to select broad or structural indicators of sentiment (e.g., \FR{مشکل} : "problem", \FR{فساد} : "corruption"), whereas humans privilege direct insults, sarcasm, and culturally salient cues (e.g., \FR{خر} : "stupid", \FR{سگ} : "dog"). In surprise or fear cases, GPT-4o often highlights neutral topical markers like \FR{نمودار} ("chart") or \FR{انسانیت} ("humanity"), while humans select emotionally charged expressions such as \FR{برگام ریخت} ("I was shocked") or \FR{هیولا} ("monster").  

% Table~\ref{tab:qualitative-gpt4t} presents representative samples comparing GPT-4-turbo and human annotations under both Predict-then-Explain (P-E) and Explain-then-Predict (E-P) settings. We observe that GPT-4-turbo frequently highlights structurally or thematically relevant words (e.g., \FR{کتابخون} "well-read", \FR{مامانت} "your mom"), whereas human annotators consistently point to affective slang or evaluative markers (e.g., \FR{خر} "tupid", \FR{گند} "mess") as decisive signals of emotion. Even in positive cases, such as texts labeled Happiness, humans emphasize intensifiers like \FR{بیش باد} ("may it increase") that convey strong cultural and emotional resonance, while the model focuses on neutral topical references such as named entities.

% Table~\ref{tab:qualitative-gpt4o} provides analogous examples for GPT-4o. Here we see a similar trend: the model tends to select broad or structural indicators of sentiment (e.g., \FR{مشکل} "problem", \FR{فساد} "corruption"), whereas humans privilege direct insults, sarcasm, and culturally salient cues (e.g., \FR{خر} "stupid", \FR{سگ} "dog"). In surprise or fear cases, GPT-4o often highlights neutral topical markers like \FR{نمودار} ("chart") or \FR{انسانیت} ("humanity"), while humans select emotionally charged expressions such as \FR{برگام ریخت} ("I was shocked") or \FR{هیولا} ("monster").  

Taken together, these qualitative cases illustrate that even the most underlying faithful GPT-family models remain misaligned with human reasoning when interpreting emotional language. The gap arises primarily from underweighting colloquial, affective, and culturally grounded expressions, features that are especially critical in low-resource and slang-rich languages like Persian.

\renewcommand{\FR}[1]{{\footnotesize\textRL{#1}}}
\begin{table*}
\small
\renewcommand{\arraystretch}{1.3}
\caption{Representative Persian samples comparing GPT-4-turbo and human-identified influential words under Predict-then-Explain (P-E) and Explain-then-Predict (E-P) settings.}
\label{tab:qualitative-gpt4t}
\centering
\begin{tabularx}{\textwidth}{@{}l c >{\raggedleft\arraybackslash}X >{\raggedleft\arraybackslash}X@{}}
\toprule
\textbf{Setting} & \textbf{Label} & \textbf{Model (Top-$k$ Words)} & \textbf{Human (Top-$k$ Words)} \\
\midrule % \midrule
\multicolumn{4}{>{\raggedleft\arraybackslash}p{\textwidth}@{}}{\FR{من خر نیستم؛ دهن که باز می‌کنی می‌فهمم که آخرین کتابی که خوندی برمی‌گرده به چهار سالگیت، تازه اونم مامانت خونده برات نه خودت، پس برای من ادای روشن‌فکرای کتاب‌خون خفن رو در نیار که واقعاً چندشم میشه!!! نصف شبی گند زدی تو اعصابم!}} \\
\midrule
\textbf{PE} & Anger & {\FR{چندشم، گند، اعصابم، نیار، خر}} & {\FR{خر، دهن، می‌فهمم، گند، اعصابم}} \\
\midrule
\textbf{EP} & Anger & {\FR{کتاب، مامانت، روشن، فکرای، کتابخون}} & {\FR{خر، دهن، می‌فهمم، گند، اعصابم}} \\
\midrule % \midrule
\multicolumn{4}{>{\raggedleft\arraybackslash}p{\textwidth}@{}}{\FR{شهر خرم آبادم ، تو اتاق انتهایی خونه مادربزرگم در حالی که پنجره بازه ... صدای جیر جیرک و شاخ و برگ درختا میاد و هوای سرشار از اکسیژن اتاق رو پر کرده ، پاهام از زیر پتو بیرونه و با نسیم خنکی چند دقیقه یک بار ماساژ داده میشه}} \\
\midrule
\textbf{PE} & Happiness & {\FR{نسیم، خنکی، پنجره، بازه، اکسیژن}} & {\FR{ماساژ، خنکی، سرشار، نسیم، اکسیژن}} \\
\midrule 
\textbf{EP} & Happiness & {\FR{خرم، آبادم، پنجره، جیرک، نسیم}} & {\FR{ماساژ، خنکی، سرشار، نسیم، اکسیژن}} \\
\midrule % \midrule
\multicolumn{4}{>{\raggedleft\arraybackslash}p{\textwidth}@{}}{\FR{\#فرشته\_حسینی ستاره این روزهای سینمای ایران، در \#جشنواره\_فیلم\_فجر هم حضور دارد. \#سیدرضامحمدی شاعر است و مقام نخست جشنواره \#شعرفجر را بهدست آورد. این دو هنرمند و شاعر اصالتا اهل افغانستان هستند. موفقیت هر دو هموطن افغانستانی در فضای سینما و ادبیات ایران خوشایند است. بیش باد.}} \\
\midrule
\textbf{PE} & Happiness & {\FR{خوشایند، موفقیت، ستاره، شاعر، افغانستان}} & {\FR{خوشایند، موفقیت، بیش باد، هموطن، ستاره}} \\
\midrule
\textbf{EP} & Happiness & {\FR{فرشته\_حسینی، جشنواره\_فیلم\_فجر، سیدرضامحمدی، شعرفجر، افغانستان}} & {\FR{خوشایند، موفقیت، بیش باد، هموطن، ستاره}} \\
\bottomrule
\end{tabularx}
\end{table*}

\renewcommand{\FR}[1]{{\footnotesize\textRL{#1}}}
\begin{table*}
\small
\renewcommand{\arraystretch}{1.3}
\caption{Representative Persian samples comparing GPT-4o and human-identified influential words under Predict-then-Explain (P-E) and Explain-then-Predict (E-P) settings.}
\label{tab:qualitative-gpt4o}
\centering
\begin{tabularx}{\textwidth}{@{}l c >{\raggedleft\arraybackslash}X >{\raggedleft\arraybackslash}X@{}}
\toprule
\textbf{Setting} & \textbf{Label} & \textbf{Model (Top-$k$ Words)} & \textbf{Human (Top-$k$ Words)} \\
\midrule % \midrule
\multicolumn{4}{>{\raggedleft\arraybackslash}p{\textwidth}@{}}{\FR{مشکل و بدبختی ایران همینجاست که تموم فسادا از زیر عمامه امثال تو میاد بیرون و برای فریب مردم گیر دادین به دو تار موی زنان... ریش شما ریشه مردم رو سوزونده... خدا خر رو شناخته که بهش شاخ نداده... این عوعو سگان شما هم نیز بگذرد... \#انسانم\_آرزوست}} \\
\midrule
\textbf{PE} & ANGER & {\FR{مشکل، بدبختی، فساده، فریب، سوزونده}} & {\FR{امثال، خر، سگ، فریب، بدبختی}} \\
\midrule
\textbf{EP} & Hatred & {\FR{مشکل، بدبختی، فسادا، عمامه، فریب}} & {\FR{امثال، خر، سگ، فریب، بدبختی}} \\
\midrule % \midrule 
\multicolumn{4}{>{\raggedleft\arraybackslash}p{\textwidth}@{}}{\FR{میتونم با نمودار و عدد و رقم بهتون نشون بدم اثر درمان دارویی روی روح و روان من باورنکردنیه. خودمم عمق داستان رو نمیدونستم تا وقتی که دو دسته دیتا از پارسال و امسال رو کنار هم گذاشتم و برگام ریخت. کاش زودتر به فکرش میفتادم.}} \\
\midrule
\textbf{PE} & Surprise & \FR{باورنکردنیه، داستان، دیتا، برگام، ریخت} & \FR{باورنکردنی، برگام، نمیدونستم، عمق، اثر} \\
\midrule
\textbf{EP} & Surprise & \FR{نمودار، درمان، روح، روان، داستان} & \FR{باورنکردنی، برگام، نمیدونستم، عمق، اثر} \\
\midrule % \midrule
\multicolumn{4}{>{\raggedleft\arraybackslash}p{\textwidth}@{}}{\FR{\#ما\_رنگین\_کمانیها از ترس از دست دادن دوستان قدیمی و یه دنیا خاطره، از ترس انگشت نما شدن و از ترس جهل شما پارانویدها فقط بازیگران بهتری نسبت به سایرین هستیم. ما هیولا نیستیم. انسانیم و انتظاری جز انسانیت نداریم. \#انسانم\_آرزوست}} \\
\midrule
\textbf{PE} & Fear & \FR{{ترس، دست، انگشت، جهل، هیولا}} & \FR{{ترس، بازیگران، هیولا، پارانویدها، نگشت نما شدن}} \\
\midrule
\textbf{EP} & Fear & \FR{{ترس، دست، دنیا، جهل، انسانیت}} & \FR{{ترس، بازیگران، هیولا، پارانویدها، نگشت نما شدن}} \\
\midrule % \midrule
\multicolumn{4}{>{\raggedleft\arraybackslash}p{\textwidth}@{}}{\FR{یادمه شبهای سرد زمستان غرب کشور از عراقیها نمیترسیدیم.از مجاهدین ضدخلق دائم در هراس بودیم چون زبان فارسی حرف میزدن وبه پشت سنگرها نفوذ میکردن وبا سیم برنده وکلنگ سربازهای سر پست نگهبانی رو میکشتن..عملیات چلچراغ شاهد این ادعاست}} \\
\midrule
\textbf{PE} & Fear & \FR{نمیترسیدیم، هراس، نفوذ، میکشتن، عملیات} & {\FR{هراس، میکشتن، مجاهدین، ضدخلق، سیم برنده}} \\
\midrule
\textbf{EP} & Fear & \FR{شبهای، سرد، زمستان، هراس، عملیات} & {\FR{هراس، میکشتن، مجاهدین، ضدخلق، سیم برنده}} \\
\bottomrule
\end{tabularx}
\end{table*}

\end{document}